\documentclass[10pt,twocolumn,letterpaper]{article}

\usepackage[pagenumbers]{cvpr}    % To produce the REVIEW version
\usepackage[accsupp]{axessibility}
\usepackage{graphicx}
\usepackage{amsmath}
\usepackage{multirow}
\usepackage{booktabs}
\usepackage{multirow}
\usepackage{amssymb}
\usepackage{booktabs}
\usepackage{colortbl}

\usepackage[pagebackref,breaklinks,colorlinks]{hyperref}

\usepackage[capitalize]{cleveref}
\crefname{section}{Sec.}{Secs.}
\Crefname{section}{Section}{Sections}
\Crefname{table}{Table}{Tables}
\crefname{table}{Tab.}{Tabs.}

\def\cvprPaperID{0002} % *** Enter the CVPR Paper ID here
\def\confName{DLGC}
\def\confYear{2024}

\begin{document}
\newcommand{\myunderbrace}[2]{\settowidth{\bracewidth}{#1#1}#1\hspace*{-1\bracewidth}\smash{\underbrace{\makebox{\phantom{#1#1}}}_{#2}}}
\title{RDPN6D: Residual-based Dense Point-wise Network for 6Dof Object Pose Estimation Based on RGB-D Images}

\author{Zong-Wei Hong \space\space\space\space Yen-Yang Hung \space\space\space\space Chu-Song Chen\\
National Taiwan University\\
{\tt\small \{r10922190, r11922a18\}@g.ntu.edu.tw, chusong@csie.ntu.edu.tw}
% \and
% Yen-Yang Hung\\
% National Taiwan University\\
}
\maketitle

%%%%%%%%% ABSTRACT
\begin{abstract}
%In this work, we present a novel method for determining the 6DoF pose of an object from a single RGB-D image. Unlike existing methods that either directly predict the object's pose or rely on sparse keypoints for pose recovery, our approach addresses this challenging task using dense correspondence, i.e., it regresses the object coordinates for each visible pixel. Our approach leverages readily available object detection methods. A re-projection mechanism is introduced to change the camera intrinsic matrix to handle cropping in RGB-D images.

In this work, we introduce a novel method for calculating the 6DoF pose of an object using a single RGB-D image. Unlike existing methods that either directly predict objects' poses or rely on sparse keypoints for pose recovery, our approach addresses this challenging task using dense correspondence, i.e., we regress the object coordinates for each visible pixel. Our method leverages existing object detection methods. We incorporate a re-projection mechanism to adjust the camera’s intrinsic matrix to accommodate cropping in RGB-D images.
%Moreover, we transform the 3D object coordinates into a residual representation, proving highly effective in reducing the output space yielding superior performance. 
Moreover, we transform the 3D object coordinates into a residual representation, which can effectively reduce the output space and yield superior performance.
%Our approach builds upon the foundation of 2D-3D dense correspondence methods that have been successful in RGB-based 6DoF pose estimation. 
%However, instead of limiting to establishing correspondences between image coordinates and object coordinates, we extend to 2D/3D (to) 3D correspondence by incorporating depth information. 
%This augmentation enables us to fully leverage the geometric constraints inherent to rigid objects, making it more accessible for the network to learn. 
We conducted extensive experiments to validate the efficacy of our approach for 6D pose estimation. 
%The results demonstrate notable improvements over the state-of-the-art methods, particularly in occlusion scenarios. 
Our approach outperforms most previous methods, especially in occlusion scenarios, and demonstrates notable improvements over the state-of-the-art methods. Our code is available on \href{https://github.com/AI-Application-and-Integration-Lab/RDPN6D}{https://github.com/AI-Application-and-Integration-Lab/RDPN6D}.
%Our end-to-end approach balances between simplicity and accuracy, achieving comparable performance to the state-of-the-art while maintaining a low computational cost.
\end{abstract}

% \begin{figure*}[t!]
% \centering
% \begin{subfigure}{1\linewidth}
%    \centering
%     \includegraphics[width=0.7\linewidth,height=3.5cm]{fig1a.pdf}
%     \caption{The pipeline of previous  state-of-the-art keypoint-based methods. \cite{he2020pvn3d, he2021ffb6d, wu2022vote}}\label{fig:image1}
% \end{subfigure}

% \bigskip
% \begin{subfigure}{\linewidth}
%   \centering
%   \includegraphics[width=0.85\linewidth %, height=4.5cm
%   ]{fig1b.pdf}
%   \caption{The pipeline of our proposed network RDPN.}\label{fig:image3}
% \end{subfigure} 
% \caption{Conventional keypoint-based methods for RGB-D 6D object pose estimation require the model to find a predefined set of sparse keypoints in 3D space. In contrast, our method predicts the 3D coordinates of each pixel on the object's surface, resulting in dense correspondences. This is achieved by combining a wider anchor with a finer-tuned residual vector. This representation eliminates the need for the model to directly produce the precise coordinates within the expansive 3D space, making it more robust and efficient. Additionally, our method can achieve improved performance in heavily occluded scenes.}
% \label{fig:example of residual representation}
% \end{figure*}
%%%%%%%%% BODY TEXT
\section{Introduction}
\label{sec:intro}
%6DoF object pose estimation is a pivotal technological advancement with wide-ranging applications in augmented reality \cite{marchand2015pose}, autonomous driving \cite{geiger2012we,chen2017multi}, and robotic manipulation \cite{tremblay2018deep, collet2011moped}. 
Estimating the pose of every object in an image is an essential task. % in computer vision. 
This paper introduces a simple but effective %method for 
6DoF object pose estimation method based on RGB-D images.
Object pose estimation is a crucial technique in many applications, such as augmented reality \cite{marchand2015pose}, autonomous driving \cite{geiger2012we,chen2017multi}, and robotic manipulation \cite{tremblay2018deep, collet2011moped}.
Traditional methods \cite{collet2011moped, lowe1999object, hinterstoisser2013model} use handcrafted features to establish 2D-3D correspondences between the image and 3D object meshes.
However, they face challenges such as sensor noise, %changing 
variable lighting conditions, and visual ambiguity.
Recent methods leverage deep learning using convolutional neural networks (CNNs) or Transformer-based \cite{amini2021t6d} architectures in RGB images to address these challenges. 
These methods typically provide %dense 
pixel-level correspondences \cite{li2019cdpn, wang2021gdr, hodan2020epos, di2021so} or predict 2D image locations for predefined 3D keypoints \cite{peng2019pvnet, amini2022yolopose}, resulting in more robust results. Nonetheless, they still face challenges such as handling textureless objects and severe occlusion.

%%\begin{figure}[t]
%  \centering
%    \includegraphics[width=0.7\linewidth]{Fig1.pdf}
%    \caption{An exemplification of our coarse-to-fine approach based on residuals. For each individual pixel situated upon an object's surface, we possess the capability to portray it through a wider anchor coupled with a finer-tuned residual vector. This specific representation eliminates the need for the model to directly produce the precise coordinates within the expansive space. }
   %\label{fig:example of residual representation}
%\end{figure}
%%%
% In pursuit of more robust performance, one avenue within deep learning is to establish dense correspondences. For instance, \cite{li2019cdpn, wang2021gdr, hodan2020epos} employ CNNs to construct pixel-wise relationships between images and object surfaces, showcasing their resilience against occlusions and clutter, Nevertheless, these approaches have certain limitations.
% First, these methods often directly regress object coordinates and assume a unimodal distribution, which can be problematic in cases involving visual ambiguities. Second, these approaches typically require the use of Perspective-Point (PnP) algorithms to estimate pose through the projected equation, potentially leading to discrepancies in scale errors between the 2D image space and the 3D space. Moreover, some geometric information of rigid objects may be lost during the process \cite{he2020pvn3d}.

To address the issues mentioned above, and because the cost of RGB-D sensors continues to decrease, recent studies such as PVN3D \cite{he2020pvn3d}, FFB6D \cite{he2021ffb6d}, RCVPose \cite{wu2022vote}, and DFTr \cite{Zhou_2023_ICCV} have employed RGB-D images for 6DoF object pose estimation.
%extended the previously mentioned RGB method into the 3D space. 
%Given the extensive output space, in contrast to the prevailing trend of establishing dense correspondences in RGB methods, their approach involves predicting predefined keypoints in the 3D space (i.e., establish 3D-3D sparse correspondences) and subsequently solving for the pose using a least-square fitting method, as shown in \cref{fig:image1}. 
%While these sparse keypoint methods exhibit strong performance, they do come with a few drawbacks. First, they involve time-consuming processes like keypoint voting or least-square fitting. Second, predicting keypoints that are heavily occluded remains a challenging task.
Given the large output space in 6DoF, existing methods based on RGB-D images predict predefined keypoints in the 3D space (i.e., establishing 3D-3D sparse correspondences). % and subsequently solve for the pose using a least-square fitting method, as illustrated in \cref{fig:image1}. 
The pose is then solved using least squares fitting methods.
Although sparse keypoint methods exhibit essential %visual 
clues that are explainable for object pose estimation, they still suffer from some drawbacks.
First, they often involve time-consuming procedures such as keypoint voting. % or least squares fitting.
Second, predicting the keypoints that are heavily occluded remains challenging. Third, they might fail to find object landmarks under viewpoint changes \cite{su2022zebrapose}.

% This stands in contrast to the predominant research trend in the RGB scheme, which primarily focuses on regressing 3D object coordinates.
% It is important to highlight that the current study, which revolves around predicting correspondences in the RGB-D scheme, has thus far been limited to the prediction of sparse keypoints. This deviation is primarily attributable to the substantial output space, for which the keypoint-based method exhibits superior performance, as demonstrated in \cite{he2020pvn3d}.

%On the other hand, recent studies in the RGB-D scheme have aimed to address real-time requirements by directly predicting object poses without time-consuming post-processing, such as DenseFusion \cite{wang2019densefusion} and ES6D \cite{mo2022es6d}. Uni6D \cite{jiang2022uni6d} and Uni6Dv2 \cite{sun2023uni6dv2} have also sought to integrate RGB feature extraction and depth feature extraction, achieving outstanding performance in terms of both accuracy and speed.

Another approach in the RGB-D scheme aims to directly predict the object poses from the feature representations generated by the neural networks, such as DenseFusion \cite{wang2019densefusion} and ES6D \cite{mo2022es6d}. %Uni6D \cite{jiang2022uni6d} and Uni6Dv2 \cite{sun2023uni6dv2} have also sought to integrate RGB feature extraction and depth feature extraction for a direct pose prediction.
Uni6D \cite{jiang2022uni6d} and Uni6Dv2 \cite{sun2023uni6dv2} have also %attempted to integrate 
integrated RGB and depth feature extraction for direct pose prediction.
The pose-deriving procedure is less explainable because the object pose is regressed directly from the feature embedding.
The object poses estimated are often less accurate using the direct pose estimation approaches.
This weakness is also observed in a related track of study, Camera Pose Estimation (aka Ego-motion Estimation) \cite{chen2022dfnet, c2f-ms-transformer2023}.
%An explanation on why the pose-regression method is less accurate 
Reasons why pose regression methods are less accurate are provided in \cite{Sattler_2019_CVPR}.

\begin{figure}[t!]
 \includegraphics[width=\linewidth]{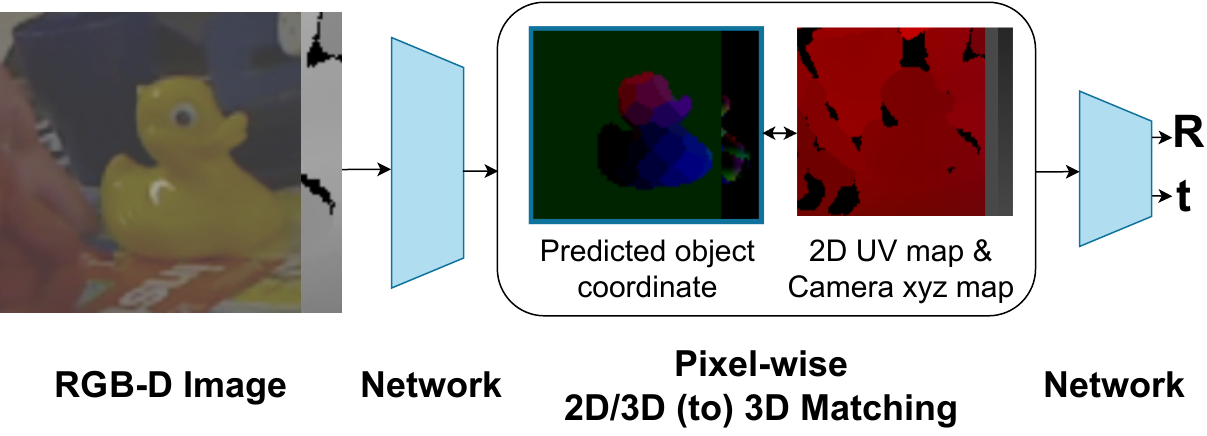}
  \caption{%\textbf{The pipeline of our proposed network RDPN.} 
  \textbf{Overview of our approach.}
  %Conventional keypoint-based methods for RGB-D 6D object pose estimation require the model to find a predefined set of sparse keypoints in 3D space. In contrast, 
  Our method predicts the 3D coordinates of each pixel on the object's surface, resulting in pixel-wise (or dense) correspondence. The object pose is then estimated based on the pixel-wise correspondence.
  }
\label{fig:method_overview}
\end{figure}

To overcome the limitations of the above-mentioned RGB-D image-based methods, we propose to use dense correspondence in 3D space to mitigate the shortcomings of sparse correspondence and pose regression.
This involves the utilization of both dense 2D-3D correspondence and 3D-3D correspondence, where the former employs the RGB-channels and the latter employs the D-channel inputs to establish the correspondences in association with the point clouds of 3D object models. 
To better combine the RGB and depth information, we modify the approach of ES6D \cite{mo2022es6d}.
Our approach generates intermediate feature representations for the RGB images and the positionally encoded depth map (camera xyz map) at first. Then, it uses the concatenated feature maps to produce the final embedding jointly.
%Besides, instead 
Instead of directly regressing the pose from the embedding, we use the embedding to bring about the dense correspondence from the joint intermediate representation.
This avoids the shortcomings of the pose regression approach, which is more %closely 
related to pose approximation than to accurate pose estimation. %via 3D structure.
Our dense-correspondence method maintains explainability and can achieve better accuracy than the sparse-correspondence method because complete 2D-3D and 3D-3D correspondence information is used. An overview is given in \cref{fig:method_overview}.

%Building upon the aforementioned observations, in this research, we expand 2D-3D dense correspondence-based methods into the 3D space by incorporating additional depth information (i.e., camera xyz map) to establish dense 2D-3D-3D correspondences. This enhancement allows us to harness the complete geometric constraint information of rigid objects. 

Furthermore, to avoid the need to predict the coordinates in a large and relatively unlimited space when finding the correspondence, we transform the coordinates of the 3D points in the object database into \emph{residual-based} representations.
A set of scattered anchor points is selected from the 3D model, and the coordinate of an object point is expressed as the residue vector for its nearest anchor point, as illustrated in \cref{fig:residual_representation}. 
The representation ensures that the prediction range is more condensed and focused.
Subsequently, %in response to real-time requirements, 
%we introduce a simple yet highly efficient framework for integrating depth and RGB information to enhance the robustness of object coordinate estimation and the prediction of 
the object pose is estimated using a neural network based on the dense correspondence, % through the utilization of these correspondences 
%instead of relying on 
which is more efficient than the time-consuming process such as PnP-RANSAC.

To evaluate our proposed residual-based dense %correspondence 
point network (RDPN) approach %comprehensively
in a broad range of scenarios, we conducted experiments using four benchmark datasets: MP6D \cite{chen2022mp6d}, YCB-Video \cite{xiang2018posecnn}, LineMOD \cite{hinterstoisser2013model}, and Occlusion LineMOD \cite{brachmann2014learning}.  Our approach generally outperforms most state-of-the-art methods on the four datasets, especially on heavily occluded datasets. Specifically, our approach achieves the best ADD-S AUC on the YCB-Video and MP6D datasets and the best ADD(-S) 0.1d on the LineMOD and Occlusion LineMOD datasets. %Additionally, we conduct comprehensive ablation studies to demonstrate the effectiveness of our design decisions.

%-------------------------------------------------------------------------

% Update the cvpr.cls to do the following automatically.
% For this citation style, keep multiple citations in numerical (not
% chronological) order, so prefer \cite{Alpher03,Alpher02,Authors14} to
% \cite{Alpher02,Alpher03,Authors14}.

\section{Related work}
%Given an input image, 
Two approaches are mainly used to infer the object poses related to the 3D models stored in the object database: approaches based on RGB images and RGB-D images.

In RGB-based approaches, SSD6D \cite{kehl2017ssd} and T6D-Direct \cite{amini2021t6d} adopt object detection networks to infer an additional output for the object pose. 
YOLOPose \cite{amini2022yolopose} uses keypoint matching to derive the rotation and translation.
%\subsection{RGB Dense Correspondences Prediction}
%Unlike keypoint-based methods, dense correspondence approaches tipically predict the 3D coordinates of corresponding object points for each visible pixel, which enhances their robustness to occlusions. Dense correspondence is commonly used in RGB images, but less so in RGB-D images due to the large output space. We review some RGB dense correspondence works below. For example, 
DPOD\cite{zakharov2019dpod} projects the 3D model points onto a 2-channel texture map and then predicts the coordinates %of each pixel 
on the corresponding texture map. Pix2Pose  \cite{park2019pix2pose} uses an encoder-decoder architecture to predict the 3D coordinates and also uses a GAN to improve stability. CDPN \cite{li2019cdpn} predicts correspondences in two steps and directly predicts translations from images instead of relying on the results of PnP-RANSAC. To address symmetric objects, EPOS \cite{hodan2020epos} discretizes the object surface into fragments and predicts a probability distribution over fragments. % per pixel. 
GDR-Net \cite{wang2021gdr} utilizes a simple yet effective 2D convolutional Patch-PnP to regress the 6D pose directly. It also introduces a symmetry-aware module from EPOS \cite{hodan2020epos} to handle symmetric objects. ZebraPose \cite{su2022zebrapose} proposes a %novel coarse to fine surface encoding technique to encode each surface vertex hierarchically, 
a hierarchical surface encoding technique and then efficiently solves the pose through the one-to-one correspondence.

Despite their excellent performance, %on RGB images, dense correspondence methods 
approaches based on RGB images are still limited by texture-less objects and severe occlusions, which can result in wrong 2D-3D correspondences. 
%Our work extends previous work into 3D space by providing additional depth information to construct 2D-3D-3D correspondences.
With the help of depth-sensing information, approaches based on RGB-D images can generally perform better.
Most RGB-D based methods %in 6Dof object pose estimation 
assume pre-segmentation of object instances in the input image.
After instance segmentation, the pose of each object is estimated individually.
Existing approaches can be divided into two main categories: Direct Pose Prediction and Keypoint-based Prediction.
The former regresses the poses of the segmented objects from the feature embedding layer directly in the image.
It is computationally more efficient, but the pose obtained relies directly on the feature embedding, which would be less accurate and lacks explainability.
The latter learns to select a set of keypoints (or proposals) for each object in the database.
Given a segmented object in the input image, the corresponding points are found and matched at first.
The object pose is then estimated based on the keypoint correspondence between the input image and the 3D models using closed-form solutions with outlier removal (e.g.,~RANSAC) or a network for pose estimation.
The approach is more explainable since the keypoints acting as proposals are learnable and we know the correspondence between the input images and the database objects.
However, the sparsity of keypoints may restrict the performance.

They are reviewed respectively in the following.

\subsection{Direct Pose Prediction}
Researches in \cite{wang2019densefusion, mo2022es6d, jiang2022uni6d, sun2023uni6dv2} opt for a direct prediction of rotation and translation. %for the sake of computational efficiency. 
% In the RGB scheme, SSD6D \cite{kehl2017ssd} and T6D-Direct \cite{amini2021t6d} adapt the object detection network to include an additional output for the object pose. In the RGB-D scheme, 
DenseFusion \cite{wang2019densefusion} introduces the densefusion module, which fuses RGB and depth features at the pixel level and predicts the object pose accordingly. The final predicted object pose is determined by selecting the pixel with the highest confidence level. ES6D \cite{mo2022es6d} presents XYZNet as a solution to mitigate the need for random memory accesses in the densefusion module to improve time efficiency. Additionally, they introduce the A(M)GPD loss, specifically designed to address the challenges posed by symmetric objects, as an improvement over the previous ADD-S loss. Experimental results demonstrate that the A(M)GPD loss yields greater efficacy in handling symmetric objects. In Unit6D \cite{jiang2022uni6d} and Unit6Dv2 \cite{sun2023uni6dv2}, the authors address the issue of projection breakdown by introducing supplementary UV data as an input, %. This breakthrough enables 
enabling a unified backbone to estimate the object pose accurately.

While direct pose prediction (or regression) methods are time-efficient, %the performance are usually degraded 
they usually have degraded performance compared to keypoint-based methods due to sensor noise.

\subsection{Keypoint-based Prediction}
Unlike direct pose prediction methods, keypoint-based approaches \cite{he2020pvn3d,he2021ffb6d, wu2022vote, Zhou_2023_ICCV} improve robustness using the projection equation.
These methods define predetermined keypoints on the object's surface and predict their positions in the image frame or camera coordinate system. The object’s pose is then computed based on these correspondences through %a time-consuming process such as 
PnP-RANSAC or least-square fitting \cite{fischler1981random,arun1987least}.

% PVNet \cite{peng2019pvnet} proposes a pixel-level voting network to predict keypoints in the image frame.  Building on this,
PVN3D \cite{he2020pvn3d} extends PVNet \cite{peng2019pvnet} to predict the keypoints in the 3D space because errors that may appear small in the projection can significantly impact the real world. FFB6D \cite{he2021ffb6d} enhances PVN3D \cite{he2020pvn3d} by incorporating bidirectional fusion modules to share information between the two modalities at an early stage. RCVPose \cite{wu2022vote} proposes a novel keypoint voting scheme that uses 1D radial voting to mitigate cumulative errors in each channel, which can significantly improve the accuracy of keypoint localization predictions. DFTr \cite{Zhou_2023_ICCV} fuses two cross-modal features using a transformer block to communicate global information between the two modalities. Moreover, instead of using the MeanShift \cite{comaniciu2002mean} algorithm to perform keypoint voting, DFTr \cite{Zhou_2023_ICCV} proposes non-iterative weighted vector-wise voting scheme to reduce computational costs.

% In addition to 3D coordinate regression, recent research has explored addressing visual ambiguities by treating the prediction of correspondences as a multi-model distribution. For instance, SurfEmb \cite{haugaard2022surfemb} employs a contractive loss to unsupervisedly train the model to learn dense correspondences between 2D images and visible object coordinates. EPro-pnp \cite{chen2022epro} uses a Monte Carlo Pose loss to penalize incorrect dense correspondence sets.

\section{Method}
Given an RGB-D image $\mathcal{I}$ and the set of 3D CAD models $\mathcal{M} = \{\mathcal{M}_i| i = 1, ..., N\}$, the primary objective of 6Dof object pose estimation is to accurately determine the rotation $\textbf{R} \in SO(3)$ and the 
translation $\textbf{t} \in \mathbb{R}^3$ that correspond to the camera coordinate system for each detected object $\mathcal{O} = \{\mathcal{O}_i| i = 1, ..., O\}$ in the image $\mathcal{I}$. 
\subsection{Overview}
Previous methods combining RGB and depth features at the pixel level require either an external instance segmentation network \cite{wang2019densefusion,mo2022es6d} or an integrated module designed for object segmentation \cite{he2020pvn3d,he2021ffb6d, jiang2022uni6d, sun2023uni6dv2}. %detection \cite{jiang2022uni6d, sun2023uni6dv2}. 
To address the demands of %real-time processing 
processing efficiency and the challenges posed by small objects, we adopt a two-step approach similar to the approaches proposed in \cite{li2019cdpn,wang2021gdr} that rely on object detection.
%However, unlike them using RGB images, our approach uses RGB-D images and how to process the depth information (represented as the camera xyz map) is a key to the successfulness for object pose estimation.
However, unlike those that use RGB images, our approach uses RGB-D images. Processing the depth information (represented as the camera xyz map) is crucial to the success of object pose estimation.

\begin{figure*}[t]
  \centering
    \includegraphics[width=0.75\linewidth]{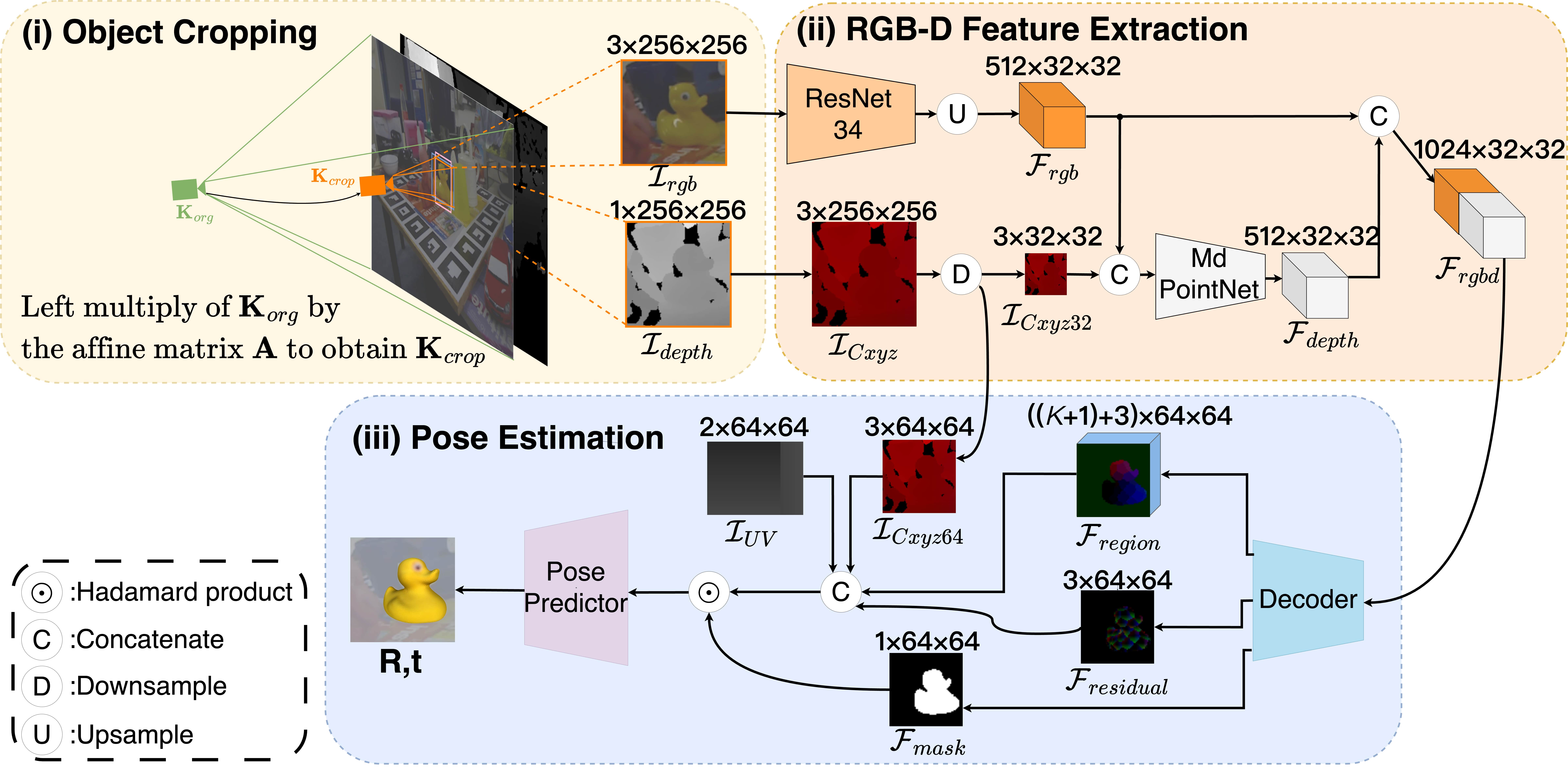}
    \caption{\textbf{Framework of RDPN.} \textbf{(i)} Starting with an RGB-D image, our initial step involves utilizing the outcomes of object detection to crop the region of interest (ROI), which results in a zoomed-in view ($\mathcal{I}_{rgb}, \mathcal{I}_{depth}$),  In order to obtain the accurate projected camera xyz map $\mathcal{I}_{C_{xyz}}$, it is necessary to adjust the original camera intrinsic $\textbf{K}_{org}$ to $\textbf{K}_{crop}$. 
    \textbf{(ii)} Once we have prepared the $\mathcal{I}_{rgb}$ and $\mathcal{I}_{C_{xyz}}$, %we can employ an RGB-D feature extractor. This 
    the RGB-D feature extractor is responsible for capturing the RGB-D fusion features $\mathcal{F}_{rgbd}$, and %feeding 
    feed them into a feature decoder to obtain both the mask ($\mathcal{F}_{mask}$) and per-pixel prediction to %regression of object coordinates. 
    the point coordinates in the 3D model of the object. This includes a ($\textit{K}+1$)-dimensional region probability ($\mathcal{F}_{region}$),  3-dimensional corresponding nearest anchors, and the residual vector ($\mathcal{F}_{residual}$). \textbf{(iii)} Finally, based on the mask and object coordinates, we utilize an image uv map  ($\mathcal{I}_{UV}$) and a downsampled camera xyz map  ($\mathcal{I}_{C_{xyz64}}$) to establish dense correspondences. These correspondences are then input into the pose predictor to regress the object pose \textbf{R} and \textbf{t}.
    } 
   \label{fig:arch}
\end{figure*}

Our proposed network, RDPN, is illustrated in \cref{fig:arch}. First, we use a readily available object detector %(e.g.,  YOLOX \cite{ge2021yolox}) 
to derive detection results. Afterward, for each detection, we crop the RGB-D image using the detected bounding box %information 
to obtain resized RGB-D images ($\mathcal{I}_{rgb} \text{ and } \mathcal{I}_{depth}$). We first use the camera intrinsic matrix to derive the camera xyz map to utilize the depth information better.
To obtain an accurate projected camera xyz map, we adjust $\textbf{K}_{org}$, the original camera intrinsic matrix, to the adjusted intrinsic matrix $\textbf{K}_{crop}$ for the cropped window, and apply it to obtain the camera xyz map ($\mathcal{I}_{C_{xyz}}$). Then, we feed $\mathcal{I}_{rgb} \text{ and } \mathcal{I}_{C_{xyz}}$ to our network. The network predicts both coarse anchors ($\mathcal{A}_{coarse}$) and residual vector ($\mathcal{F}_{residual}$) for each visible pixel of the object, as well as a mask ($\mathcal{F}_{mask}$) for removing background pixels. Leveraging these predictions, in conjunction with the 2D UV map ($\mathcal{I}_{UV}$) and down-sampled camera xyz map ($\mathcal{I}_{Cxyz64}$), our model %directly 
 predicts the object pose jointly using %2D-3D-3D 
 the 2D-3D and 3D-3D dense correspondences established.

%------------------------------------------------------------------------
\begin{figure}[t]
\centering
    \includegraphics[width=0.75\linewidth]{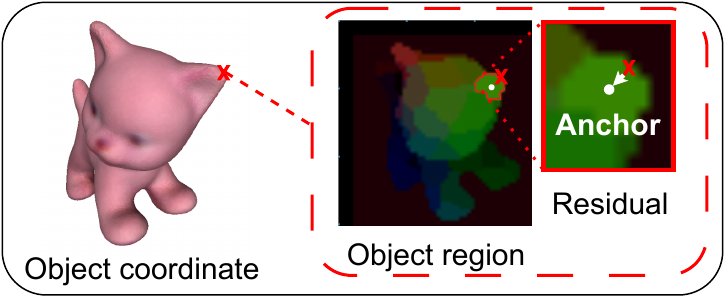}
\caption{
 \textbf{Our purposed residual representation.} 
  We use distributedly located anchor points and a fine-level residual vector (to the nearest anchor) to map each point on the object's surface. % to the nearest pre-selected anchor in the region. 
  This eliminates the need for the network to directly predict the exact coordinates where the range is extensively large, and makes the correspondence prediction more robust. % and efficient.
 }
\label{fig:residual_representation}
\end{figure}

\subsection{Handling Cropping in RGB-D Images}
To obtain the camera xyz map $\mathcal{I}_{C_{xyz}}$, we can utilize \cref{eq:projection equation}, which describes the projection of a point on the object's surface ($X,Y,Z$) into the point of the image plane ($u, v$) through the camera intrinsic matrix $\textbf{K}$ after being rotated with rotation matrix $\textbf{R}$ and translated by $\textbf{t}$:

\begin{equation}
\begin{aligned}
\begin{bmatrix}
u \\ v \\ 1
\end{bmatrix}
&= \frac{\textbf{K}} {d} \cdot
\begin{pmatrix}
\textbf{R}
\begin{bmatrix}
X \\ Y \\ Z 
\end{bmatrix}
+ \textbf{t}
\end{pmatrix}
\end{aligned}
,
\label{eq:projection equation}
\end{equation}
where $d$ is a scale factor.

However, once we have acquired the object bounding boxes, we must crop and resize the images to obtain the Regions of Interest (RoI). This can be seen as a sub-image obtained by applying an affine transformation to the original image or modifying the intrinsic matrix of the original image. When transforming the depth image into a camera xyz map using the original intrinsic matrix, the %projection 
projected location will be %onto 
incorrect %locations 
within the camera. % coordinate system. 
Therefore, it is necessary to adjust the original camera intrinsic matrix $\textbf{K}_{org}$ by left multiplying it by an affine matrix $\textbf{A}$ (as illustrated in the upper left part of \cref{fig:arch}), which is a $3\times3$ matrix with the last row $[0~0~1]$. This gives us $\textbf{K}_{crop} = \textbf{A} \textbf{K}_{org}$, which ensures accurate projection for the RoI.

%In addition to achieving accurate projections, adjusting the intrinsic matrix also addresses the problem of diverse distribution of object positions between the training and testing datasets. Specifically, if we directly use Korg\textbf{K}_{org}, the object 3D points could appear in %different 
%a wide range of locations in the image (e.g., the upper-left or lower-right). However, after adjusting the intrinsic matrix to Kcrop\textbf{K}_{crop}, the object will always be %fixed 
%located in the center of the image, making the subsequent network easier to learn from the consistently spread coordinates of the object location.

In addition to achieving accurate projections, adjusting the intrinsic matrix addresses the diverse distribution of object positions between the training and testing datasets. Specifically, if we directly use $\textbf{K}_{org}$, the object 3D points could appear in a wide range of locations in the image, such as the upper-left or lower-right. However, after adjusting the intrinsic matrix to $\textbf{K}_{crop}$, the object will always be located in the center of the image, making it easier for the subsequent networks to robustly learn from uniformly distributed coordinates for correspondence finding.

\label{sec:method DZI}

\subsection{Residual Representation}
\label{sec:method residual}
%Previous dense-correspondence methods 
A possible approach to dense correspondence is to predict the 3D object coordinates of all object pixels in a single step. % \cite{li2019cdpn,wang2021gdr}. 
It is efficient in this way but requires the model to identify each mesh point within the image and store its coordinates in an extensively large range of the object coordinate system. 
%Learning to predict them is a challenging task because the large output space and can lead to suboptimal performance for objects with complex shapes or symmetric features \cite{he2020pvn3d}. 
Learning to predict them is challenging because of the large output space, which can lead to suboptimal performance for objects with complex shapes or symmetric features \cite{he2020pvn3d}. 
Additionally, symmetric features could cause ambiguities in the mapping, as there may be multiple points on the object %equally 
similar to the given pixel in an image.

%To counteract this limitation, 
To address this concern, we propose a novel residual-based strategy that decouples the prediction of the object coordinates into the coarse and fine parts. For the coarse part, we first establish multiple anchors for the object $\mathcal{M}$ using Farthest Point Sampling (FPS) \cite{623193}. FPS effectively subdivides the object's surface into distinct subregions. Specifically, we can establish the anchor set $\mathcal{A}_{i}$ for the given object $\mathcal{M}_i$ as follows:
\begin{equation}
    \mathcal{A}_i = \{\mathcal{A}_{i}^k| \mathcal{A}_{i}^k \in FPS(\mathcal{M}_i, K) , \forall k=1, ..., K \},
\end{equation} 
where %FPS(⋅)FPS(\cdot) is farthest point sampling and 
$K$ %denotes the region number.
is the number of anchors.
%For each object in the CAD database, its anchor points can divide the object's 3D point cloud into separate regions according to the nearest neighbors to the anchors.
Each object in the 3D model database can be divided into separate regions based on its anchor points, determined by nearest-neighbor grouping. %to the anchors. %in the object’s 3D point cloud.
For each point on the object's surface, we then use these anchors as reference points and compute the %offset 
residual vector (i.e., fine part) to its nearest anchor, as illustrated in \cref{fig:residual_representation}. 
%Through the utilization of a residual representation, we can reformulate the projection equation as \cref{eq:residual projection equation} in the following:
By utilizing a residual representation, we can reformulate the projection equation as \cref{eq:residual projection equation} below:

\begin{equation}
\begin{aligned}
\begin{bmatrix}
u \\ v \\ 1
\end{bmatrix}
&= \frac{\textbf{K}} {d} \cdot
\begin{pmatrix}
\textbf{R}
\begin{bmatrix}
X \\ Y \\ Z 
\end{bmatrix}
+ \textbf{t}
\end{pmatrix}
\\[10pt]
 &= \frac{\textbf{K}} {d} \cdot
\begin{pmatrix}
\textbf{R}
\smash{\underbrace{
\begin{bmatrix}
R_x \\ R_y \\ R_z
\end{bmatrix}
}_{coarse}}
+
\textbf{R}
\smash{\underbrace{
\begin{bmatrix}
r_x \\ r_y \\ r_z 
\end{bmatrix}
}_{fine}}
+ \textbf{t}
\end{pmatrix}
\end{aligned}
\label{eq:residual projection equation}
\end{equation}
\newline
\newline
\newline
where $(R_x, R_y, R_z) \in \mathcal{A}_i$ represents the closest anchor point of the given point $(X, Y, Z)$ on the object's surface and $(r_x, r_y, r_z) = (X - R_x, Y - R_y, Z - R_z)$ denotes the residual of the point. This residual projection %equation establishes 
enables a %comprehensive dense 
condensed and uniform range to represent the correspondence relationship between the camera frame and the object coordinate system. %and allows us to avoid using the %large 
%diverse output space of original object coordinates.

% \begin{equation}
% \begin{aligned}
% \begin{bmatrix}
% u \\ v \\ 1
% \end{bmatrix}
%  = \frac{\textbf{K}} {d} \cdot
% \begin{pmatrix}
% \textbf{R}
% \smash{\underbrace{
% \begin{bmatrix}
% R_x^i \\ R_y^i \\ R_z^i
% \end{bmatrix}
% }_{Coarse}}
% +
% \smash{\underbrace{
% \textbf{R}
% \begin{bmatrix}
% r_x^i \\ r_y^i \\ r_z^i 
% \end{bmatrix}
% }_{fine}}
% + \textbf{t}
% \end{pmatrix}
% ,
% \end{aligned}
% \label{eq:residual projection equation}
% \end{equation}

\subsection{RDPN}
Our proposed RDPN takes as input a cropped RGB image $\mathcal{I}_{rgb}$, a camera xyz map $\mathcal{I}_{C_{xyz}}$, and the predicted goal coarse part and fine part for the visible object pixels. The network is designed using an encoder-decoder architecture to predict the object coordinates for each pixel. The predicted object coordinates, the 2D UV map ($\mathcal{I}_{UV}$), and the downsampled camera xyz map ($\mathcal{I}_{C_{xyz64}}$) are then used to compute the object pose via a simple but effective network.

\noindent\textbf{RGB-D Feature Encoder}. 
Our approach to fusing the information from the two modalities draws inspiration from ES6D \cite{mo2022es6d}. Nevertheless, a critical distinction between their method and ours is that we do not concatenate  $\mathcal{I}_{rgb}$ and $\mathcal{I}_{C_{xyz}}$ initially and then feed them into CNNs to extract local features. %We avoid this approach because of the potential for background noise to hinder effective fusion of the two modalities within our detection-based approach.
We avoid this approach because background noise may hinder our detection-based approach's effectiveness for fusion of the two modalities.

Instead, we extract local texture features ($\mathcal{F}_{rgb}$) from the RGB image $\mathcal{I}_{rgb}$ using CNNs. We then combine these features with the downsampled camera xyz map ($\mathcal{I}_{Cxyz32}$) and input them into a PointNet-like CNNs architecture, using $1\times1$ convolutions to compress both position and texture information for individual pixels to get the spatial features ($\mathcal{F}_{depth}$). We empirically concatenate the two features at a $32\times32$ resolution to leverage the depth information %better
efficiently. Finally, we obtain the global RGB-D fusion feature ($\mathcal{F}_{rgbd}$) by concatenating the $\mathcal{F}_{rgb}$ and $\mathcal{F}_{depth}$, as shown in the upper right part of \cref{fig:arch}.

\noindent\textbf{RGB-D Feature Decoder}.
The residual representation of the 3D model object coordinates is predicted by the RGB-D feature decoder, as illustrated in the lower part of \cref{fig:arch}.
%The RGB-D feature decoder predicts the database's object coordinates in the form of residual representation depicted in Sec.~????????????????????????????????????????????????????????????????????????\ref{sec:method residual} based on the RGB-D fusion feature Frgbd\mathcal{F}_{rgbd}. 
 %guide the network to 
To utilize the dense correspondences accurately, we also let the decoder network predict the visible object mask $\mathcal{F}_{mask}$ and guide it with ground truth $\hat{\mathcal{F}}_{mask}$ by applying $\mathcal{L}_1$ loss:
\begin{equation}
    \mathcal{L}_{mask} = || \mathcal{F}_{mask} - \hat{\mathcal{F}}_{mask}||_1.
\end{equation}
For the coarse part in the residual representation, we approach it as a \textbf{classification problem}, opting to choose the anchor with the highest probability for each pixel. 
Consider the anchor set of the object $i$, $\mathcal{A}_i = \{\mathcal{A}_{i}^k| k=1, ..., K \}$.
The object points can be divided into regions by these $K$ anchors according to nearest-neighbor grouping.
This defines the ground-truth labels of the point in a region: if the region is associated with anchor $k$, every point inside this region is labeled by a one-hot vector with the $k$-th element $1$ and the others $0$.
The one-hot vector is of length $K+1$, where the $(K+1)$-th element denotes the background.
Hence, every spatial site of the RGB-D fusion feature $\mathcal{F}_{rgbd}$ can be classified as one of the classes in $\{1,\cdots, (K+1)\}$ after decoding. We use $\mathcal{\hat{F}}_{region}$ to represent the classification labels thus obtained.
%The ground-truth coarse region ˆFregion\mathcal{\hat{F}}_{region} can be derived from Ai\mathcal{A}_i by grouping the object points whose nearest anchor is Ai\mathcal{A}_i. 
%In this region, the RGB-D fusion feature Frgbd\mathcal{F}_{rgbd} should be classified as 
%Thus, 
Hence, we can supervise the predicted $\mathcal{F}_{region}$ with cross-entropy loss, formulate as:
\begin{equation}
    \mathcal{L}_{coarse} = Cross\_Entropy(\mathcal{F}_{mask} \odot \mathcal{F}_{region}, \mathcal{\hat{F}}_{region} ),
\end{equation}
where $\odot$ denotes the Hadamard product, % and note that the region number is K+1K + 1 due to the background channel. %,
%with the background channel set in the first channel.
$\mathcal{F}_{region}$ records the probability of being the $k$-th class ($k=1\cdots(K+1)$) for every spatial site.
%To infer the coarse correspondence locations, we require the anchor sites in addition to the class probabilities. 
To infer coarse correspondence locations, we need anchor positions in addition to class (or anchor label) probabilities.
We employ the anchor coordinate $\mathcal{A}_i^k=(R_x, R_y, R_z)$ as feature representation with $k$ the most probable class. %That is, if some position (u,v)(u,v) is classified as the kk-th class, the kk-th anchor's coordinate Aki=(Rx,Ry,Rz)\mathcal{A}_i^k=(R_x, R_y, R_z) is then recorded in the position. 
This yields an additional $3$ channels. An illustration can be found in the lower part of \cref{fig:arch}.

%Subsequently, the coarse part is prepared as follows:
%\begin{equation}
%    \mathcal{A}_{coarse} = \{\mathcal{A}_{coarse}^{u,v} | \mathcal{A}_i^k, where\textit{ } k \in \arg \max \mathcal{F}_{region}^{u,v}\},
%\end{equation}
%where Fu,vregion\mathcal{F}_{region}^{u,v} denotes the region probability for the pixel (u,v)(u,v).
%Note that to ensure stable convergence of the network, we also apply an L1\mathcal{L}_1 loss (called Lregion_background\mathcal{L}_{region\_background}) to the background channel of Fregion\mathcal{F}_{region} with the ˆFmask\hat{\mathcal{F}}_{mask}.
% :
% \begin{equation}
%     \mathcal{L}_{region\_background} = || \mathcal{F}_{region_0} - \hat{\mathcal{F}}_{mask}||_1,
% \end{equation}
% where Fregion0\mathcal{F}_{region_0}  denotes the background channel of Fregion\mathcal{F}_{region}.

For the fine part in the residual representation, we approach it as a \textbf{regression problem}. To generate the residual $\mathcal{F}_{fine}$, we employ a straightforward approach by applying an $\mathcal{L}_1$ loss to the ground truth residual $\hat{\mathcal{F}}_{fine}$: 
\begin{equation}
\begin{split}
    &\mathcal{L}_{fine} = || \mathcal{F}_{mask} \odot (\mathcal{F}_{fine} - \hat{\mathcal{F}}_{fine})||_1.
\end{split}
\end{equation}

\noindent\textbf{Pose Predictor}.
%To predict object's pose through using those dense correspondences effectively, as opposed to keypoint-based approaches that relied on least-squares fitting, we adopt the methodology presented in \cite{wang2021gdr}.
%Specifically, the pose predictor (PP) is a straightforward 2D CNNs  takes into account the dense visible correspondences. 
To effectively predict an object’s pose using dense correspondences, as opposed to keypoint-based methods \cite{he2020pvn3d, he2021ffb6d, Zhou_2023_ICCV} relying on least-squares fitting, we adopt the methodology presented in \cite{wang2021gdr}. The pose predictor (PP) is a simple 2D CNN based on the dense visible correspondences.

%To obtain the dense visible correspondences, we stack the predicted object coordinates Fcoarse\mathcal{F}_{coarse}  and Ffine\mathcal{F}_{fine} with  FUV\mathcal{F}_{UV} and FCxyz64\mathcal{F}_{C_{xyz64}}. Then, we remove unrelated correspondences (i.e., background correspondences) by performing pixel-wise multiplication with Fmask\mathcal{F}_{mask}, that is, the 2D-3D-3D dense correspondences are obtained by

% \begin{equation}
% \begin{split}
%    \textbf{R}, \textbf{t} = PP( &\mathcal{F}_{coarse}, \mathcal{F}_{fine}, \mathcal{F}_{mask}, \\
%     &\mathcal{F}_{region}, \mathcal{F}_{UV}, \mathcal{F}_{Cxyz64}),
% \end{split}
% \end{equation}

%\begin{equation}
%\begin{split}
% \mathcal{F}_{2D\_3D\_3D} = \mathcal{F}_{mask} \odot \textit{Concat}(&\mathcal{F}_{coarse}, \mathcal{F}_{fine}, %\\&\mathcal{F}_{Cxyz64}, \mathcal{F}_{UV}, \mathcal{F}_{region}),
% \end{split}
%\end{equation}
%and the object's pose is obtained by
%\begin{equation}
%\begin{split}
%   \textbf{R}, \textbf{t} = \textit{PP}( \mathcal{F}_{2D-3D\rightarrow3D}).
%\end{split}
%\end{equation}
%This configuration allows the regression of the final pose. 

The overall loss function can be summarized as %$\mathcal{L}_{RDPN} = %\mathcal{L}_{region\_background} + 
\begin{equation}
\mathcal{L}_{RDPN}=\mathcal{L}_{coarse} + \mathcal{L}_{fine} + \mathcal{L}_{mask} + \mathcal{L}_{Pose}, 
\end{equation}
where $\mathcal{L}_{Pose}$ is a disentangled 6D pose loss detailed in \cite{wang2021gdr}, employed to %oversee 
monitor the estimation of the pose \textbf{R} and \textbf{t}.

\begin{table*}[htb!]
    \centering
    \arrayrulecolor{black}
    \resizebox{2\columnwidth}{!}{
    \begin{tabular}{l|c|c|c|c|c|c|c|c|c|c|c} 
    \hline
    ~ & \multicolumn{3}{c|}{RGB Approaches} & \multicolumn{8}{c}{RGB-D Approaches}\\
    
    \hline
    \arrayrulecolor[rgb]{0,0,0}
                           Method  & PVNet \cite{peng2019pvnet} & CDPN \cite{li2019cdpn} & DPODv2 \cite{shugurov2021dpodv2} & PointFusion \cite{xu2018pointfusion} & DenseFusion \cite{wang2019densefusion} & G2L-Net \cite{chen2020g2l} & PVN3D \cite{he2020pvn3d} & FFB6D \cite{he2021ffb6d}& RCVPose \cite{wu2022vote} & DFTr \cite{Zhou_2023_ICCV} & RDPN (Ours)\\ 
                           ~ & (CVPR' 19) & (ICCV' 19) &(TPAMI' 21)  & (CVPR' 18) & (CVPR' 19) & (CVPR' 20)  & (CVPR' 20)  & (CVPR' 21)  & (ECCV' 22) & (ICCV' 23) &\\
    % ~ & \cite{xu2018pointfusion} & \cite{wang2019densefusion} &   & \cite{he2020pvn3d} & \cite{he2021ffb6d} & \cite{wu2022vote} & \cite{Zhou_2023_ICCV}\\
    \arrayrulecolor{black}\hline

    ape              & 43.6 & 64.4 & \textbf{100.0}                      &  70.4           &  92.3    &  96.8           &  97.3   & 98.4    & 99.2  & 98.6     & 99.7       \\ 
    
    benchvise           & 99.9 & 97.8 & \textbf{100.0}                  &  80.7           &  93.2    &  96.1           &  99.7   & \textbf{100.0}   & 99.6  & \textbf{100.0}    & \textbf{100.0}         \\   
    
    camera                & 86.9 & 91.7&  \textbf{100.0}               &  60.8           &  94.4    &  98.2           &  99.6   & 99.9    & 99.7  & \textbf{100.0}    & 99.9  \\
    
    can                   & 95.5 & 95.9&  \textbf{100.0}               &  61.1           &  93.1    &  98.0           &  99.5   & 99.8    & 99.0  & \textbf{100.0}    & \textbf{100.0}                                          \\ 

    cat                   & 79.3 & 83.8&  \textbf{100.0}               &  79.1           &  96.5     & 99.2           &  99.8   & 99.9    & 99.4  & \textbf{100.0}    & \textbf{100.0}                                             \\ 
    
    driller               & 96.4 & 96.2&  \textbf{100.0}               &  47.3           &  87.0    &  99.8           &  99.3   & \textbf{100.0}   & 99.7  & \textbf{100.0}    & \textbf{100.0}                                             \\ 
    
    duck                & 52.6 & 66.8&    \textbf{100.0}               &  63.0           &  92.3    &  97.7           &  98.2   & 98.4    & 99.4  & 99.1     & \textbf{100.0}                                             \\ 
     
    \textbf{eggbox}          & 99.2 & 99.7&   99.04                   &  99.9           &  99.8    &  \textbf{100.0}          &  99.8   & \textbf{100.0}   & 98.7  & \textbf{100.0}    & \textbf{100.0} \\
    
    \textbf{glue}          & 95.7 & 99.6&   98.03                     &  99.3           &  \textbf{100.0}    & \textbf{100.0}           & \textbf{100.0}  & \textbf{100.0}   & 99.7  & \textbf{100.0}    & \textbf{100.0} \\
    
    holepuncher      & 82.0 & 85.8&   99.03                   &  71.8           &  92.1    &  99.0           &  99.9   & 99.8    & 99.8  & \textbf{100.0}    & \textbf{100.0}\\
    
    iron          & 98.9 & 97.9&   \textbf{100.0}                      &  83.2           &  97.0    &  99.3           &  99.7   & 99.9    & 99.9  &  99.9    & \textbf{100.0}  \\
    
    lamp           & 99.3 & 97.9&  \textbf{100.0}                      &  62.3           &  95.3    &  99.5           &  99.8   & 99.9    & 99.2  & \textbf{100.0}    & \textbf{100.0}\\
   
    phone        & 92.4 & 90.8&   \textbf{100.0}                       &  78.8           &  92.8    &  98.9           &  99.5   & 99.7    & 99.1  & 99.6     & \textbf{100.0}\\
    \hline
    Avg (13)    & 86.3 & 89.9&   99.70                        &  73.7           &  94.3    &  98.7           &  99.4   & 99.7    & 99.43  & 99.8     & \textbf{99.97}                                              \\ 
    
    \hline
    \end{tabular}
    }
    \caption{Quantitative evaluation of 6D Pose ADD(-S) 0.1d on the LineMOD dataset. Symmetric objects are in bold.}
    \label{tab: lm results}
    \arrayrulecolor{black}
    \end{table*}
\begin{table*}[htb!]
    \centering
    \arrayrulecolor{black}
    \resizebox{2\columnwidth}{!}{
    \begin{tabular}{l|c|c|c|c|c|c|c|c|c} 
    %\hline
    %~ & \multicolumn{3}{c|}{RGB} & \multicolumn{8}{c}{RGB-D}\\
    
    \hline
    \arrayrulecolor[rgb]{0,0,0}
                            Method      & PoseCNN \cite{xiang2018posecnn} & Hybridpose \cite{song2020hybridpose} & PVN3D  \cite{he2020pvn3d}  & FFB6D \cite{he2021ffb6d}  & RCVPose \cite{wu2022vote} & Uni6D \cite{jiang2022uni6d} & Uni6Dv2 \cite{sun2023uni6dv2} &DFTr \cite{Zhou_2023_ICCV}  & RDPN (Ours) \\ 
    ~       & (RSS' 18)  & (CVPR' 18)  & (CVPR' 20) & (CVPR' 21) & (ECCV' 22) & (CVPR' 22)  & (AISTATS' 23)  & (ICCV' 23)   & ~   \\
    \hline

    ape             & 9.6   & 20.9 & 33.9  & 47.2 & 60.3  & 33.0 & 44.3   & \underline{64.1}  &  \textbf{64.6}   \\
    
    can             & 45.2  & 75.3 & 88.6  & 85.2 & 92.5  & 51.0  & 53.3   &  \underline{96.1}  &  \textbf{97.0}   \\

    cat             & 0.9   & 24.9 & 39.1  & 45.7 & 50.2  & 4.6  & 16.7   &  \underline{52.2}  &  \textbf{54.8}  \\
    
    driller         & 41.4  & 70.2 & 78.4  & 81.4 & 78.2  & 58.4  & 63.0   & \textbf{95.8}  &   \underline{93.1}  \\  
    
    duck            & 19.6  & 27.9 & 41.9  & 53.9 & 52.1  & 34.8  & 38.1   & \textbf{72.3}  &   \underline{68.8}  \\
     
    \textbf{eggbox}         & 22.0  & 52.4 & \underline{80.9}  & 70.2 & \textbf{81.2}  & 1.7  & 4.6   & 75.3  &   78.1  \\
    
    \textbf{glue}           & 38.5  & 53.8 & 68.1  & 60.1 & 72.1  & 30.2  & 40.3   &  \underline{79.3}  &  \textbf{83.5}  \\
    
    holepuncher     & 22.1  & 54.2 & 74.7  & 85.9 & 75.2  & 32.1  & 50.9   &  \underline{86.8}  &  \textbf{96.1}   \\
    \hline
    Avg (8)         & 24.9  & 47.5 & 63.2  & 66.2 & 70.2  & 30.7  & 40.2   &  \underline{77.7}  & \textbf{79.5}   \\
    \hline
    \end{tabular}
    }
    \caption{Quantitative evaluation of 6D Pose ADD(-S) 0.1d on the Occlusion-LineMOD dataset. Symmetric objects are in bold.}
    \label{tab: lmo results}
    \arrayrulecolor{black}
    \end{table*}

\section{Experiments}
In this section, we present the experimental results of our RDPN method and comparisons to the other approaches.

\subsection{Implementation Details}
We implemented our %experiments using the PyTorch framework \cite{NEURIPS2019_9015} 
approach with PyTorch framework on a single RTX 3090 GPU. 
%For object detection, we %follow \cite{liu2022gdrnpp_bop} and 
We employ YOLOX-x \cite{ge2021yolox} for object detection. %, which utilizes stronger data augmentation and the Ranger optimizer \cite{liu2019variance, yong2020gradient}.
The number of anchors is selected as $K=32$.
%For RDPN, we trained all networks using the Ranger optimizer with a batch size of 24 and an initial learning rate of 1e-4. This learning rate was gradually reduced using a cosine schedule at 72\% of the training process. The network architecture details can be found in the supplementary material.
To address the issue of varying object sizes, we adopt Dynamic Zoom-In (DZI) to scale ground truth bounding boxes, as suggested in \cite{li2019cdpn,wang2021gdr}.
More implementation details can be found in the supplementary material.

\begin{table*}[htb!]
\centering
\arrayrulecolor{black}
\resizebox{2\columnwidth}{!}{
\begin{tabular}{l|cc|cc|cc|cc|cc|cc|cc|cc} 
\hline
\arrayrulecolor[rgb]{0,0,0}
                          & \multicolumn{2}{c|}{PVN3D \cite{he2020pvn3d}} & \multicolumn{2}{c|}{FFB6D  \cite{he2021ffb6d}} & \multicolumn{2}{c|}{RCVPose \cite{wu2022vote}} & \multicolumn{2}{c|}{ES6D \cite{mo2022es6d}} & \multicolumn{2}{c|}{Uni6D \cite{jiang2022uni6d}} & \multicolumn{2}{c|}{Uni6Dv2 \cite{sun2023uni6dv2}} & \multicolumn{2}{c}{DFTr \cite{Zhou_2023_ICCV}} & \multicolumn{2}{c}{RDPN (Ours)} \\ 
\arrayrulecolor{black}\hline
                          & ADD-S & ADD(-S)             & ADD-S & ADD(-S)             & ADD-S & ADD(-S)               & ADD-S & ADD(-S)            & ADD-S  & ADD(-S)            & ADD-S & ADD(-S)               & ADD-S  & ADD(-S)                & ADD-S  & ADD(-S)                                     \\ 
\hline
002\_master\_chef\_can    & 96.0 & 80.5                & 96.3 & 80.6                & 95.7 & \textbf{93.6}                  & 96.5 & 73.0               & 95.4 & 70.2                & 96.0 & 74.2                  &  97.0    &  92.3                            &  \textbf{99.8}  & 78.7                                              \\ 

003\_cracker\_box         & 96.1 & 94.8                & 96.3 & 94.6                & 97.2 & \textbf{95.7}         & 95.3 & 94.0               & 91.8 & 85.2                & 96.0 & 94.2                  &  95.9    &  93.9                            &  \textbf{98.3}  & 88.3                                                   \\ 

004\_sugar\_box           & 97.4 & 96.3                & 97.6 & 96.6                & 97.6 & 97.2                  & 97.9 & 97.3               & 96.4 & 94.5                & 97.6 & 96.6                  &  97.1    &  95.5                            &  \textbf{100.0}  & \textbf{99.9}                                                   \\ 

005\_tomato\_soup\_can    & 96.2 & 88.5                & 95.6 & 89.6                & 98.2 & 94.7                  & 97.3 & 90.4               & 95.8 & 85.4                & 96.1 & 86.6                  &  95.6   &  92.6                            &  \textbf{98.4}  & \textbf{97.1}                                                 \\ 

006\_mustard\_bottle      & 97.5 & 96.2                & 97.8 & 97.0                & 97.9 & 97.2                  & 98.2 & \textbf{97.9}      & 95.4 & 91.7                & 97.8 & 96.7                  &  97.6    &  96.3                            &  \textbf{99.7}  & 97.0                                                   \\ 

007\_tuna\_fish\_can      & 96.0 & 89.3                & 96.8 & 88.9                & 98.2 & 96.4                  & 97.4 & 93.7               & 95.2 & 79.0                & 96.3 & 76.0                  &  97.3    &  94.5                            &  \textbf{100.0}  & \textbf{99.2}                                                   \\ 

008\_pudding\_box         & 97.1 & 95.7                & 97.1 & 94.6                & \textbf{97.7} & \textbf{97.1}                  & 96.5 & 93.4               & 94.1 & 89.8                & 96.6 & 94.7                  &  97.4    &  95.7          & 95.0  & 89.3                                               \\ 

009\_gelatin\_box         & 97.7 & 96.1                & 98.1 & 96.9                & 97.7 & 96.5                  & 97.7 & 96.5               & 97.4 & 96.2                & 98.0 & 97.0                  &  97.6    &  96.3                            & \textbf{100.0} & \textbf{100.0}                                                 \\ 

010\_potted\_meat\_can    & 93.3 & 88.6                & 94.7 & 88.1                & \textbf{97.9} & 90.2                  & 92.5 & 84.6               & 93.0 & 89.6                & 95.7 & 91.9   &  95.9    &  \textbf{92.1}                         & \textbf{97.9} & 90.1                                                    \\ 

011\_banana               & 96.6 & 93.7                & 97.2 & 94.9                & 97.9 & 96.7                  & 97.9 & 85.8               & 96.4 & 93.0                & 98.0 & \textbf{96.9}         &  97.1    &  95.0                            & \textbf{100.0} & 96.6                                                 \\ 

019\_pitcher\_base        & 97.4 & 96.5                & 97.6 & 96.9                & 96.2 & 95.7                  & 97.8 & 97.7               & 96.2 & 94.2                & 97.5 & 96.9                  &  96.0    &  93.1                            & \textbf{100.0} & \textbf{98.7}                                                 \\ 

021\_bleach\_cleanser     & 96.0 & 93.2                & 96.8 & 94.8                & \textbf{99.2} & \textbf{97.8}                  & 96.3 & 92.8               & 95.2 & 91.1                & 97.0 & 95.3    &  96.8    &  94.9                        & 98.3 & 89.8                                                 \\ 

\textbf{024\_bowl}                & 90.2 & 90.2                & 96.3 & 96.3                & 95.2 & 94.9                  & 96.4 & 96.4               & 95.5 & 95.5                &96.8 & 96.8    &   \textbf{96.9}    &   \textbf{96.9}                        & 87.8 & 87.8                                          \\ 

025\_mug                  & 97.6 & 95.4                & 97.3 & 94.2                & 98.4 & 96.3                  & 97.3 & 95.0               & 96.6 & 93.0                & 97.7 & 96.3          &  97.6    &  94.9                           & \textbf{100.0} & \textbf{97.9}                                                   \\ 

035\_power\_drill         & 96.7 & 95.1                & 97.2 & 95.9                & 96.2 & 95.4                  & 97.2 & 96.3               & 94.7 & 91.1                & 97.6 & 96.8                   &  96.9    &  95.2                           & \textbf{99.9} & \textbf{97.8}                                                    \\ 

\textbf{036\_wood\_block}         & 90.4 & 90.4                & 92.6 & 92.6                & 89.1 & 89.3                  & 94.4 & 94.4               & 94.3 & 94.3                & 96.1 & 96.1                  &  96.2   &  96.2                            & \textbf{96.7} & \textbf{96.7}                                                    \\ 

037\_scissors             & 96.7 & 92.7                & \textbf{97.7} & \textbf{95.7}                & 96.2 & 94.7                  & 87.1 & 61.5               & 87.6 & 79.6                & 95.0 & 90.3   &  97.2    &  93.3                         & 96.6 & 89.3                                                    \\ 

040\_large\_marker        & 96.7 & 91.8                & 96.6 & 89.1                & 95.9 & 92.4                  & 97.8 & 90.6               & 96.7 & 92.8                & 97.0 & 93.1                     &  96.9    &  92.7                         & \textbf{99.9} & \textbf{93.9}                                                   \\ 

\textbf{051\_large\_clamp}        & 93.6 & 93.6                & 96.8 & 96.8                & 95.2 & 96.4                  & 61.0 & 61.0               & 95.9 & 95.9                & 97.0 & 97.0                     &  96.3    &  96.3                         & \textbf{98.9} & \textbf{98.9}                                                    \\ 

\textbf{052\_extra\_large\_clamp} & 88.4 & 88.4                & 96.0 & 96.0                & 94.7 & 94.7                  & 59.6 & 59.6               & 95.8 & 95.8                & 96.5 & 96.5                     &  96.4    &  96.4                         & \textbf{98.6} & \textbf{98.6}                                                    \\ 

\textbf{061\_foam\_brick}         & 96.8 & 96.8                & 97.3 & 97.3                & 95.7 & 95.7                  & 96.6 & 96.6               & 96.1 & 96.1                & 97.4 & 97.4                      &  97.3    &  97.3                        & \textbf{100.0} & \textbf{100.0}                                                   \\ 
\hline
Avg (21)                   & 95.5 & 91.8                & 96.6 & 92.7                & 96.6 & \textbf{95.2}                  & 93.6 & 89.0               & 95.2 & 88.8                & 96.8 & 91.5            &  96.7    &  94.4                  &  \textbf{98.4} & 94.6                                           \\
\hline
\end{tabular}
}
\caption{Quantitative results of 6D Pose (ADD-S AUC and ADD(-S) AUC) on the YCB-Video dataset. Symmetric objects are in bold.}
\label{tab: ycbv results}
\arrayrulecolor{black}
\end{table*}

\begin{table*}[htb!]
    \centering
    \arrayrulecolor{black}
    \resizebox{2\columnwidth}{!}{
    \begin{tabular}{l|c|c|c|c|c|c|c|c|c|c} 
    \hline
    \arrayrulecolor[rgb]{0,0,0}
                              % & Hodan \cite{hodavn2015detection} & PointFusion \cite{xu2018pointfusion}  & DCF  \cite{liang2018deep}   & DF (per-pixel) & MaskedFusion \cite{pereira2020maskedfusion} &  G2L-Net \cite{chen2020g2l}  & PVN3D  \cite{he2020pvn3d}  & FFB6D \cite{he2021ffb6d}   & DFTr \cite{Zhou_2023_ICCV}  & RDPN (Ours) \\ 
                               & Hodan & PointFusion & DCF    & DF (per-pixel) & MaskedFusion  &  G2L-Net  & PVN3D   & FFB6D    & DFTr  & RDPN \\
                               &   \cite{hodavn2015detection} &\cite{xu2018pointfusion} &  \cite{liang2018deep}&  \cite{wang2019densefusion}  & \cite{pereira2020maskedfusion} & \cite{chen2020g2l} & \cite{he2020pvn3d} &  \cite{he2021ffb6d} &  \cite{Zhou_2023_ICCV} & (Ours) \\
    \arrayrulecolor{black}\hline

    \hline
    Avg (20)                  & 74.20              & 75.54                & 75.93               & 79.84             & 79.38               & 83.51              &  85.42          &  86.29             &  93.01      & \textbf{95.90} \\
    \hline
    \end{tabular}
    }
    \caption{Quantitative evaluation of 6D Pose (ADD-S AUC)  on the MP6D dataset. DF (per-pixel) denotes DenseFusion (per-pixel). Note that all objects are symmetric.}
    \label{tab: mp6d results}
    \arrayrulecolor{black}
    \end{table*}

\subsection{Benchmark Datasets}
\noindent\textbf{LineMOD}\cite{hinterstoisser2013model} contains sequences of 13 objects with mild occlusion, cluttered scenes, texture-less objects, and variations in lighting conditions, presenting challenges for accurate pose estimation in real-world scenarios.
Our approach follows the protocol established in \cite{peng2019pvnet, xiang2018posecnn, wu2022vote, wang2021gdr, di2021so}, which employs the standard 15\%/85\% training/testing split. %In addition to this split, during the training phase, 
During training, we leverage further 1k rendered images for each object, %similar to the %methodology 
as described in the settings %described 
in \cite{li2019cdpn, wang2021gdr, di2021so}.

\noindent\textbf{Occlusion LineMOD}\cite{brachmann2014learning} is an extension of LineMOD that contains challenging test images of partially occluded objects. It involves training on LineMOD images and testing Occlusion LineMOD to assess the robustness of handling heavily occluded objects.

\noindent\textbf{YCB-Video}\cite{xiang2018posecnn} is an extensive dataset including 130K key frames from 92 videos featuring 21 objects with varying lighting conditions and occlusions. Following \cite{he2020pvn3d, xiang2018posecnn, he2021ffb6d}, we split the dataset into 80 training videos and select 2,949 keyframes from the remaining 12 videos for testing.

\noindent\textbf{MP6D} \cite{chen2022mp6d} is a challenging dataset for the  6D pose estimation of metal parts in industrial environments. It contains 77 RGB-D video segments with simultaneous multi-target, occluded, and illumination changes. 
All objects are \textit{symmetric}, textureless, and of complex shape, high reflectivity, and uniform color, which makes them difficult to distinguish.
We follow \cite{chen2022mp6d, Zhou_2023_ICCV} to split the training and testing sets.

For Occlusion LineMOD and YCB-V, we also follow \cite{wang2021gdr, di2021so} that utilizes synthetic data with physically-based rendering \cite{hodavn2020bop} for training.

\noindent\subsection{Metrics}
We follow previous works \cite{xiang2018posecnn, he2021ffb6d, jiang2022uni6d, hodavn2016evaluation} and use the common average distance metrics ADD-S and ADD(-S). First, let us consider the ADD metric that evaluates the average of pairwise points transformed by the predicted pose [\textbf{R}, \textbf{t}] and the ground truth pose $[\textbf{R}^*, \textbf{t}^*]$
\begin{equation}
    \text{ADD} = \frac{1}{m} \sum_{x \in \mathcal{O}} ||(\textbf{R}x + \textbf{t} ) - (\textbf{R}^*x + \textbf{t}^*) ||,
\end{equation}
where $x$ denotes the vertex sampled from the given 3D model $\mathcal{O}$, and mm denotes the total number of vertices. 

The ADD-S metric measures the closest point distance between two point clouds, making it more suitable for evaluating symmetric objects.
\begin{equation}
    \text{ADD-S} = \frac{1}{m} \sum_{x_1 \in \mathcal{O}} \min_{x_2 \in \mathcal{O}} ||(\textbf{R}x_1 + \textbf{t} ) - (\textbf{R}^*x_2 + \textbf{t}^*) ||.
\end{equation}
ADD(-S)\cite{hinterstoisser2013model} calculates ADD for non-symmetric objects and ADD-S for symmetric objects, respectively.

For Linemod and Occlusion LineMOD datasets, we follow \cite{jiang2022uni6d, peng2019pvnet} to report ADD(-S) 0.1d. This metric measures the accuracy of distances less than 10\% of the objects' diameter. 
For the YCB-Video dataset, we follow \cite{he2021ffb6d, Zhou_2023_ICCV} to report the area under the accuracy-threshold curve computed by varying the distance threshold ADD-S AUC and ADD(-S) AUC. 
For MP6D, we report ADD-S AUC, following \cite{chen2022mp6d, he2021ffb6d, jiang2022uni6d,xiang2018posecnn} since all objects are symmetric.

\subsection{Comparison with State-of-the-Art Methods}
\noindent\textbf{Results on LineMOD \&  Occlusion LineMOD}.
The quantitative results of our proposed RDPN are presented in \cref{tab: lm results} and \cref{tab: lmo results} for the LineMOD and Occlusion LineMOD datasets, respectively. Our method achieves state-of-the-art performance on both datasets. Particularly, in occlusion scenes, our approach outperforms most previous methods extensively, %state-of-the-art method DFTr~\cite{Zhou_2023_ICCV} by 1.8\%. This improvement demonstrates 
demonstrating %the effectiveness of using dense correspondences and residual representation.
its effectiveness.

\noindent\textbf{Results on YCB-V}.
Results of our proposed RDPN on the YCB-Video dataset are shown in \cref{tab: ycbv results}. Our approach outperforms the previous state-of-the-art approaches, achieving a superior ADD-S AUC metric by 1.7\%. It also demonstrates competitive performance in the ADD(-S) AUC without time-consuming post-processing.

\noindent\textbf{Results on MP6D}.
The results of our RDPN on the MP6D dataset are shown in \cref{tab: mp6d results}. Again, our method outperforms the previous state-of-the-art methods in the ADD-S AUC metric, demonstrating its generalizability across different datasets. Compared to other RGB-D based solutions, RDPN is more robust towards objects with texture-less or heavy reflective surfaces and severe occlusions.

\begin{figure}[h]
  \centering
    \includegraphics[width=\linewidth]{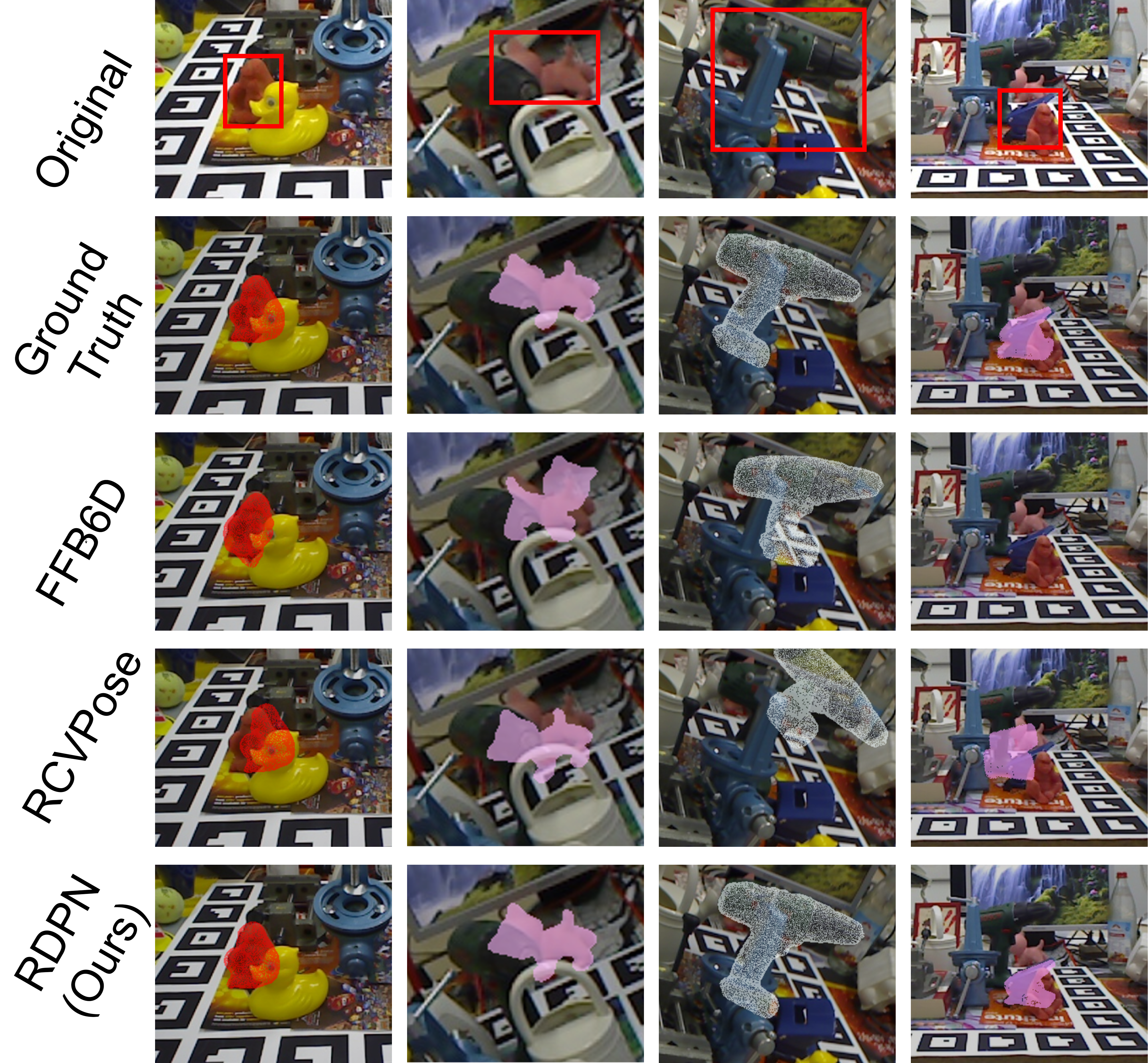}
    \caption{\textbf{Qualitative results on  Occlusion LineMOD.} The images are rendered by projecting the 3D object model onto the image plane using the estimated pose. Our two-step dense correspondence method can accurately capture the object and predict its pose, even under heavy occlusion. This contrasts previous keypoint-based methods, which often struggle in such scenarios.}
   \label{fig: vis}
\end{figure}

\noindent\textbf{Qualitative Results}.
Furthermore, to illustrate the robustness of our RDPN in handling occlusion, we present some estimation results on the Occlusion LineMOD dataset. As shown in \cref{fig: vis}, RDPN %consistently outperforms state-of-the-art methods \cite{he2021ffb6d, wu2022vote}, demonstrating 
demonstates superior performance and robustness across various occlusion scenarios.

\subsection{Ablation Study on Occlusion LineMOD}

\noindent\textbf{Effectiveness of the residual representation} is shown in \cref{tab: ablation residual}. 
%Note that for predicting the original object coordinate, we still make the network predict the region probability Fregion\mathcal{F}_{region}. 
We compare the approaches using the orginal and our anchor+residual representations for predicting the object coordinates.
As the table shows, it is clear that residual representations outperform the original representation, highlighting the advantage of reducing the output space.

\begin{table}[h]
\centering
\resizebox{1\linewidth}{!}{
\begin{tabular}{|l|ccc|}
\hline
  \multirow{2}{*}{Object Coordinate}                        & \multicolumn{3}{c|}{Average (8)}      \\ \cline{2-4}
    &  \multicolumn{1}{c|}{ADD(-S) 0.02d} & \multicolumn{1}{c|}{ADD(-S) 0.05d} & \multicolumn{1}{c|}{ADD(-S) 0.1d} 
\\ 
\hline
  Original       & \multicolumn{1}{c|}{7.7}          & \multicolumn{1}{c|}{50.2} & \multicolumn{1}{c|}{78.5} \\ 
  Residual Representation        & \multicolumn{1}{c|}{\textbf{9.1}}          & \multicolumn{1}{c|}{\textbf{51.3}} & \multicolumn{1}{c|}{\textbf{79.5}} \\  
\hline
\end{tabular}}
\caption{Ablation on the effectiveness of residual representation.}
\label{tab: ablation residual}
\end{table}

\begin{table}[h]
\centering
\resizebox{1\linewidth}{!}{
\begin{tabular}{|c|c|ccc|}
\hline
2D-3D  &   3D-3D                        & \multicolumn{3}{c|}{Average (8)}      \\ \cline{3-5}
Correspondences  & Correspondences  &  \multicolumn{1}{c|}{ADD(-S) 0.02d} & \multicolumn{1}{c|}{ADD(-S) 0.05d} & \multicolumn{1}{c|}{ADD(-S) 0.1d} 
\\ 
\hline
v               & -        & \multicolumn{1}{c|}{4.6} & \multicolumn{1}{c|}{35.3} & \multicolumn{1}{c|}{67.3} \\ 
-               & v        & \multicolumn{1}{c|}{\textbf{9.4}}          & \multicolumn{1}{c|}{50.9} & \multicolumn{1}{c|}{78.4} \\  

v               & v        & \multicolumn{1}{c|}{9.1}          & \multicolumn{1}{c|}{\textbf{51.3}} & \multicolumn{1}{c|}{\textbf{79.5}} \\  
\hline
\end{tabular}}
\caption{Ablation study on the effectiveness of different component of dense correspondences.}
\vspace{-7px}
\label{tab: ablation dense correspondences}
\end{table}

\begin{table}[h!]
\centering
\resizebox{0.9\linewidth}{!}{
\begin{tabular}{|c|ccc|}
\hline
Intrinsic                        & \multicolumn{3}{c|}{Average (8)}      \\ \cline{2-4}
Parameter   &  \multicolumn{1}{c|}{ADD(-S) 0.02d} & \multicolumn{1}{c|}{ADD(-S) 0.05d} & \multicolumn{1}{c|}{ADD(-S) 0.1d} 
\\ 
\hline
$\textbf{K}_{org}$                      & \multicolumn{1}{c|}{7.1} & \multicolumn{1}{c|}{41.2} & \multicolumn{1}{c|}{72.4} \\ 
$\textbf{K}_{crop}$                     & \multicolumn{1}{c|}{\textbf{9.1}}          & \multicolumn{1}{c|}{\textbf{51.3}} & \multicolumn{1}{c|}{\textbf{79.5}} \\  
\hline
\end{tabular}}
\caption{Ablation study on the effectiveness of adjusting the intrinsic parameter.}
\vspace{-15px}
\label{tab: ablation K}
\end{table}

% The effectiveness of residual representation is demonstrated in \cref{fig:ablation study} (b). For a more in-depth exploration, we also converted the original object coordinates to the residual representation within the object coordinate system. To elaborate, the residual of the model's output corresponds to the natural fine part as described in \cref{eq:residual projection equation}. It is evident that both residual representations outperform the original representation when evaluating using the ADD-S metric. However, when transitioning to the ADD(-S) metric, the performance of the camera coordinate residual representation appears to degrade. We believe that this may be due to the challenging nature of precisely regressing the camera residual using the rotation matrix.

\noindent\textbf{Effectiveness of the Dense Correspondence components}
%Comparisons of the components of dense correspondence 
is compared and presented in \cref{tab: ablation dense correspondences}. We conducted experiments under three different configurations.

Initially, when the pose predictor relies solely on dense correspondence with the 2D image UV map $\mathcal{I}_{UV}$ to predict the object coordinates (i.e., 2D-3D correspondences), the performance is inferior to the previously employed direct regression methods.
However, replacing $\mathcal{I}_{UV}$ with the camera xyz map $\mathcal{I}_{C_{xyz64}}$ (i.e., 3D-3D correspondences) significantly improved all ADD(-S) metrics. This observation is intuitive, as the pose predictor can effectively predict the object's pose without considering the camera intrinsic matrix \textbf{K} through \cref{eq:projection equation}. 
Finally, utilizing dense correspondence with both $\mathcal{I}_{UV}$ and $\mathcal{I}_{C_{xyz}}$, along with the predicted object coordinates (2D/3D to 3D correspondences), further enhances the performance. % at thresholds 0.05d and 0.1d (+0.4\% and +1.1\%, respectively). 
We attribute that the sensor noise in collected depth images can be rectified by incorporating $\mathcal{I}_{UV}$, when both $\mathcal{I}_{UV}$ and $\mathcal{I}_{C_{xyz64}}$ are available.

\noindent\textbf{Effectiveness of adjusting intrinsic $\textbf{K}_{org}$}
%The effectiveness of adjusting intrinsic Korg\textbf{K}_{org} 
is shown in \cref{tab: ablation K}, which reveals %that %after adjusting Korg\textbf{K}_{org} to 
better performance %is obtained 
when using $\textbf{K}_{crop}$ to achieve accurate projection. This suggests that the intrinsic $\textbf{K}$ adjustment step is necessary in our method.

\begin{figure}[h]
  \centering
    \includegraphics[width=0.9\linewidth]{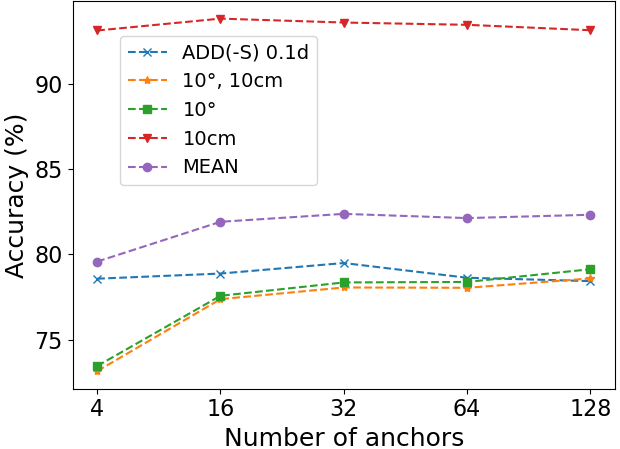}
    \caption{%Ablation study on the effectiveness of different numbers of anchors. 
    Ablation study on the number of anchors. The 10$^{\circ}$, 10 cm metric measures whether the rotation and the translation error is less than 10$^{\circ}$ and 10 cm, respectively.}
   \label{fig: ablation region}
\end{figure}

\noindent\textbf{Effectiveness of different number of anchors} is shown in
\cref{fig: ablation region}. %illustrates the effect of different numbers of anchors. 
We observe a noticeable improvement when increasing the numbers of anchors from 4 to 16, with the best performance achieved at 32 anchors. However, the performance starts to decline after using 32 anchors. We believe that as the number of anchors increases, the output space of the fine part is reduced, while the output space of the coarse part inevitably increases. Therefore, there is a trade-off between the number of anchors and overall performance.

\section{Limitation}
% object detection 66ms
% 29ms for 8 obkects
% 10ms for single objecys
Our method demonstrates remarkable speed, achieving pose estimation for 8 objects in a 640x480 image within a mere 29ms/frame. However, the overall processing time is currently constrained by the object detection stage, which requires 66ms/frame using the YOLOX-x \cite{ge2021yolox} detector. To further enhance the efficiency of our method, we plan to explore the development of a unified network for object detection and pose estimation in future work.

\section{Conclusion}
%This paper introduces 
We introduce a novel Residual-based Dense Point-wise Network (RDPN) for precise and efficient object pose estimation from RGB-D images. By employing an intrinsic adjustment and a residual representation for object coordinates, RDPN effectively condenses and concentrates the output range, eliminating the need to predict coordinates within a vast and relatively unbounded space. %Furthermore, by leveraging 
Leveraging both 2D-3D and 3D-3D dense correspondences, our method achieves state-of-the-art performance on %most 
public benchmarks with high efficacy. % while maintaining %real-time 
%efficient processing capability. %In addition to developing a unified network, we also have plans 
%In the future, we plan to tackle the challenge of handling unseen objects in the future.

\textbf{Acknowledgement} This work was supported in part by the National Science and Technology Council, Taiwan under Grant NSTC 112-2221-E-002-182-MY3 and 112-2634-F-002-005, and also under Grant UR2205 of Delta-NTU joint R\&D center. We thank to National Center for High-performance Computing (NCHC) of National Applied Research Laboratories (NARLabs) in Taiwan for providing computational and storage resources.

% \usepackage{colortbl}

%%%%%%%%% REFERENCES
\newpage
{\small
\bibliographystyle{ieee_fullname}
\bibliography{egbib}
}

\end{document}

\title{Supplementary Material - RDPN6D: Residual-based Dense Point-wise Network for 6Dof Object Pose Estimation Based on RGB-D Images}

\author{Zong-Wei Hong \space\space\space\space Yen-Yang Hung \space\space\space\space Chu-Song Chen\\
National Taiwan University\\
{\tt\small \{r10922190, r11922a18\}@ntu.edu.tw, chusong@csie.ntu.edu.tw}
% \and
% Yen-Yang Hung\\
% National Taiwan University\\
}
\maketitle

\renewcommand\thesection{\Alph{section}}
\renewcommand\thesubsection{\thesection.\arabic{subsection}}
%%%%%%%%% BODY TEXT
\section{More Implementation Details}
More implementation details are provided in this section.

\subsection{Network Architecture}
\begin{figure*}
    \centering
    \resizebox{1.5\columnwidth}{!}{
    \includegraphics{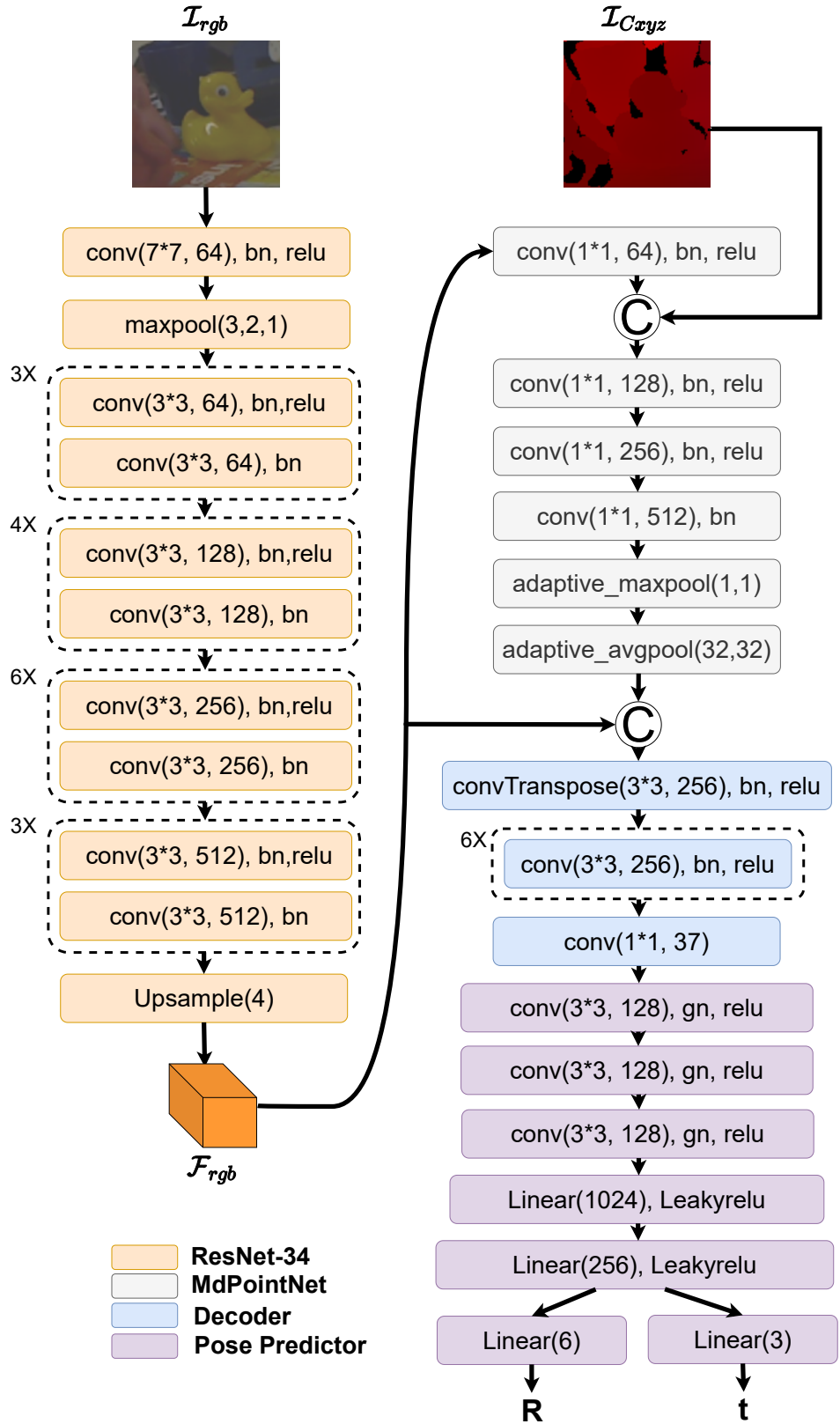}
    }
    \caption{\textbf{The detailed architecture of our proposed RDPN framework.}}
    \label{fig:arch_details}
\end{figure*}

The detailed architecture of the proposed RDPN is shown in \cref{fig:arch_details}. In this figure, $\textit{conv}(n*n, c)$ denotes a 2D convolution with kernel size $n$ and output channel $c$. $bn$ denotes batch normalization, $relu$ denotes ReLU activation, $\textit{Upsample}(s)$ denotes 2D upsampling with scale factor $s$. and $\textit{maxpool}(k,s,p)$ denotes 2D max pooling with kernel size $k$, stride $s$, and padding $p$, respectively. The output of $\textit{adaptive avgpool}(h, w)$ or $\textit{adaptive maxpool}(h, w)$ is of size $h * w$ for any input size. $\textit{convTranspose}(n*n, c)$ denotes a 2D transposed convolution with kernel size $n$ and output channel $c$. $gn$ denotes group normalization \cite{wu2018group}, $Leakyrelu$ denotes LeakyReLU activation, and $\textit{Linear}(c)$ denotes a fully connected layer with output channel $c$. 

%For the output of rotation, 
To represent rotations, we adopt the %representation 
solution proposed in \cite{zhou2019continuity} to address the issue of rotation discontinuity, which results in a 6-dimensional output.

\subsection{Training Parameters}

For RDPN, all networks were trained using the Ranger optimizer \cite{liu2019variance, zhang2019lookahead} with a batch size of 24 and an initial learning rate of 1e-4. This learning rate was gradually reduced using a cosine schedule \cite{loshchilov2016sgdr} at 72\% of the training process. 

\subsection{Training Enhancements}

 We employ two strategies to enhance the model’s ability to handle objects of varying sizes. First, we dynamically adjust the receptive field of the $\mathcal{F}_{residual}$  based on the size of the corresponding tight 3D bounding box of the CAD model. This allows the model to focus more effectively on objects of different scales.

Second, we adopt the Dynamic Zoom-In technique proposed in \cite{li2019cdpn, wang2021gdr} to alleviate the impact of varying object sizes further. During training, we randomly shift the center and scale of the ground-truth bounding boxes by a ratio of 25\%. Subsequently, we zoom in the input Regions of Interest (RoIs) with a ratio of $r$ = 1.5 while maintaining their original aspect ratio. This ensures that the area containing the object occupies approximately half of the RoIs. This dynamic zooming approach effectively normalizes the object size distribution and improves the model's generalization ability across different object sizes.
\begin{table}[ht]
\centering
\resizebox{\linewidth}{!}{
\begin{tabular}{|l|c|c|c|c|}
\hline
   \textbf{} & DenseFusion \cite{wang2019densefusion} & FFB6D \cite{he2021ffb6d} & ES6D \cite{mo2022es6d} & RDPN (Ours) \\ \hline
   ADD-S & 93.2 & 95.0  & 93.6 & \textbf{95.4}  \\ \hline
   ADD(-S)  & 86.1 & 91.3 & 89.0 & \textbf{91.5}  \\ \hline
   % RDPN (Ours) & CIR \cite{lipson2022coupled} & TBD & TBD & TBD & TBD & TBD & TBD & TBD & TBD \\ \hline
\end{tabular}}
\caption{\textbf{The YCB-V results with PoseCNN input.}}
\label{tab: posecnn results}
\end{table}

\begin{table*}
\centering
\resizebox{1\linewidth}{!}{
\begin{tabular}{|l|c|c|c|c|c|c|c|c|c|}
\hline
  \textbf{Method} & Refinement & LM-O & T-LESS & TUD-L & YCB-V & ITODD & HB & IC-BIN  & Avg(7) \\ \hline
   RCVPose 3D\_SingleModel\_VIVO\_PBR \cite{wu2022keypoint} (3DV' 22) & ICP  & 0.729 & 0.708 & \textbf{0.966} & 0.843 & 0.536 & 0.863 & \textbf{0.733} & 0.768\\ \hline
   SurfEmb-PBR-RGBD \cite{haugaard2022surfemb} (CVPR' 22)  & custom & 0.760 & 0.828 & 0.854 & 0.799 & 0.538 & 0.866 & 0.659  & 0.758 \\ \hline
   ZebraPoseSAT-EffnetB4\_refined \cite{su2022zebrapose} (CVPR' 22) & CIR \cite{lipson2022coupled} (CVPR' 22)   & \underline{0.780} & \textbf{0.862} & 0.956 & \textbf{0.899}  & \textbf{0.618} & \textbf{0.921} & 0.654 & \textbf{0.813}\\ \hline
   RADet+PFA-MixPBR-RGBD \cite{hai2023rigidity} (CVPR' 23)& PFA \cite{hu2022perspective} (ECCV' 22) & \textbf{0.797} & \underline{0.850} & \underline{0.960} & \underline{0.888} & 0.469 & 0.869 & 0.676 & 0.787\\ \hline
   %RDPN (Ours) & ICP    & \underline{0.961} & 0.893 & 0.692  & 0.848 \\ 
   RDPN (Ours) & CIR \cite{lipson2022coupled} (CVPR' 22)  & 0.776 & 0.768 & 0.957 & 0.883 & \underline{0.575} & \underline{0.907} & \underline{0.720}  & \underline{0.798}\\ \hline
\end{tabular}}
\caption{\textbf{Average Recall on the BOP Core datasets.}}
\label{tab: bop results}
\end{table*}

\begin{table*}[thb!]
    \centering
    \arrayrulecolor{black}
    \resizebox{2\columnwidth}{!}{
    \begin{tabular}{l|c|c|c|c|c|c|c|c|c|c} 
    \hline
    \arrayrulecolor[rgb]{0,0,0}
                         Method    & Hodan & PointFusion & DCF    & DF (per-pixel) & MaskedFusion  &  G2L-Net  & PVN3D   & FFB6D    & DFTr  & RDPN \\
                               &   \cite{hodavn2015detection} &\cite{xu2018pointfusion} &  \cite{liang2018deep}&  \cite{wang2019densefusion}  & \cite{pereira2020maskedfusion} & \cite{chen2020g2l} & \cite{he2020pvn3d} &  \cite{he2021ffb6d} &  \cite{Zhou_2023_ICCV} & (Ours) \\
   
    \hline
    Obj\_01                    & 83.42                 & 84.33                 & 86.06                  & 89.35                & 88.95                 & 89.51                   &  90.28               &  93.28               &  95.44              & \textbf{99.58}                                                      \\ 
    
    Obj\_02                    & 80.23                 & 81.01                 & 85.36                  & 87.78                & 89.19                 & 89.03                   &  91.88               &  92.83               &  96.51              & \textbf{99.19}                                                    \\ 
    
    Obj\_03                    & 65.78                 & 64.74                 & 65.33                  & 72.45                & 70.03                 & 74.93                   &  76.67               &  79.51               &  84.93              & \textbf{93.87}                                                       \\ 
    
    Obj\_04                    & 70.56                 & 72.50                 & 73.95                  & 77.98                & 74.68                 & 85.39                   &  88.13               &  84.98               &  92.02              & \textbf{96.36}                                              \\ 

    Obj\_05                    & 69.78                 & 68.96                 & 67.19                  & 71.23                & 75.69                 & 72.13                   &  73.46               &  76.33               &  86.24              & \textbf{95.30}                                                 \\ 
    
    Obj\_06                    & 72.36                 & 70.66                 & 71.65                  & 75.34                & 78.31                 & 85.08                   &  87.16               &  83.98               &  96.10              & \textbf{96.30}                                                 \\ 
    
    Obj\_07                    & 80.79                 & 81.12                 & 82.07                  & 88.63                & 85.25                 & 89.09                   &  94.81               &  94.94               &  97.51              & \textbf{99.33}                                                 \\ 
    
    Obj\_08                    & 80.71                 & 81.37                 & 82.39                  & 84.78                & 85.38                 & 90.10                   &  93.76               &  89.76               &  96.75              & \textbf{99.23}                                                  \\ 
    
    Obj\_09                    & 69.80                 & 65.98                 & 68.27                  & 73.67                & 75.46                 & 79.91                   &  82.71               &  81.25               &  91.23              & \textbf{95.00}                                                   \\ 

    Obj\_10                    & 75.32                 & 77.19                 & 79.10                  & 80.54                & 77.62                 & 86.03                   &  86.16               &  88.92               &  94.98              & \textbf{98.34}                                                       \\ 

    Obj\_11                    & 72.56                 & 71.98                 & 70.96                  & 79.65                & 75.91                 & 82.01                   &  81.21               &  84.87               &  92.36              & \textbf{92.55}                                                     \\ 
    
    Obj\_12                    & 74.13                 & 76.32                 & 77.03                  & 78.88                & 76.98                 & 77.93                   &  79.00               &  84.82               &  89.99              & \textbf{95.89}                                                   \\ 
    
    Obj\_13                    & 78.63                 & 77.05                 & 75.15                  & 80.12                & 80.58                 & 85.38                   &  86.69               &  85.42               &  95.04              & \textbf{95.80}                                           \\ 
    
    Obj\_14                    & 76.89                 & 77.90                 & 76.98                  & 80.89                & 81.15                 & 84.54                   &  87.06               &  87.99               &  94.13              & \textbf{94.87}                                                      \\ 
    
    Obj\_15                    & 64.53                 & 67.36                 & 66.23                  & 68.45              & 66.30                 & 72.92                  &  74.17              &  75.01              &  86.97            & \textbf{88.90}                                                       \\ 
    
    Obj\_16                    & 69.88                 & 72.28                 & 73.08                  & 75.81                & 73.86                 & 79.38                   &  81.35               &  83.95               &  92.14              & \textbf{94.93}                                                         \\ 
    
    Obj\_17                    & 77.42                 & 85.93                 & 84.68                  & 89.16                & 88.11                 & 92.08                   &  93.47               &  93.19               &  94.25              & \textbf{98.83}                                                    \\ 

    Obj\_18                    & 75.63                 & 81.46                 & 80.91                  & 83.23                & 85.94                 & 88.13                   &  87.57               &  91.73               &  94.69              & \textbf{98.13}                                                    \\ 
    
    Obj\_19                    & 72.89                 & 76.82                 & 78.07                  & 81.98                & 79.37                 & 85.31                   &  88.82               &  87.28               &  95.03              & \textbf{96.13}                                                   \\ 
    
    Obj\_20                    & 72.65                 & 75.91                 & 74.20                  & 76.59                & 78.93                 & 81.41                   &  88.10               &  85.75               &  \textbf{93.92}          & 89.56                                                     \\ 
    \hline
    Avg (20)                  & 74.20                 & 75.54                 & 75.93                  & 79.84                & 79.38                 & 83.51                   &  85.42               &  86.29               &  93.01              & \textbf{95.90}                                                    \\
    \hline
    \end{tabular}
    }
    \caption{\textbf{Quantitative evaluation of 6D Pose ADD-S AUC on the MP6D Dataset for each object.} Note that all objects are symmetric.}
    \label{tab: mp6d more results}
    \arrayrulecolor{black}
    \end{table*}

\begin{table}[tb!]
\centering
\resizebox{1\columnwidth}{!}{
\begin{tabular}{c|c|c|c|c} 
\hline
Method          & Pre-process                              & Network & Post-process & Network + Post-process \\
\hline
DenseFusion\cite{wang2019densefusion}     & \emph{IS} & 50         & 11              & 61           \\ 

FFB6D\cite{he2021ffb6d}           & -                                            & 42         & 65              & 107                                                               \\ 

ES6D\cite{mo2022es6d}    &        \emph{IS} & 6        & -               & 6                                                              \\ 

Uni6D*\cite{jiang2022uni6d}   & -                                            & 39         & -               & 39                                                                \\ 

Uni6Dv2*\cite{sun2023uni6dv2} & -                                            & 47         & -               & 47                                                                \\ 

RCVPose\cite{wu2022vote}         & -                                            & 50         & -               & 50                                                                \\ 

RDPN (Ours)           & \emph{OD}                                           & 20         & -               & 20                                                             \\
\hline
\end{tabular}
}
\caption{\textbf{Time Costs (in milliseconds per frame) on the YCB-Video Dataset.} \emph{IS} represents Instance Segmentation, and \emph{OD} represents Object Detection. %indicates the methods that have not released their source code, and we show the speed values directly from their respective papers.
(*) stands for methods whose source codes have not been released, and we %demonstrate 
report their speeds directly from their respective papers.
}
\label{tab: time}
\arrayrulecolor{black}
\end{table}

\section{More Results}
This section presents detailed evaluations of RDPN on the MP6D, YCB-Video datasets, and the BOP challenge  \cite{hodavn2020bop}.

\subsection{Quantitative Results under the same detections on the YCB-V Dataset}
To comprehensively assess the effectiveness of RDPN, we compare it with several baseline methods while ensuring a fair comparison. However, it is essential to note that while other methods utilize segmentation masks or built-in detection techniques, RDPN incorporates detection preprocessing specifically designed for RGBD images. Therefore, we adopt PoseCNN's \cite{xiang2018posecnn} RoI results for RDPN and segmentations for other methods to maintain consistency and impartiality. Despite this disparity in detection pipelines, RDPN exhibits robust accuracy, as evidenced in \cref{tab: posecnn results}. This finding underscores its efficacy even when operating under different detection paradigms.

\subsection{Quantitative Results on the BOP challenge}
\cref{tab: bop results} presents the average recall for the BOP challenge, a comprehensive benchmark for rigid body pose estimation encompassing seven diverse datasets. This benchmark has yet to reach saturation, indicating its suitability for evaluating the generalizability of pose estimation models. We evaluate RDPN on this challenge and compare its performance with published works.

\subsection{Quantitative Results on the MP6D Dataset}
The results of ADD-S AUC of each object on the MP6D
dataset are shown in \cref{tab: mp6d more results}.

\subsection{Time Costs Comparison on YCB-Video Dataset}
The time costs comparison on YCB-Video dataset are shown in \cref{tab: time}.

\subsection{Visualization on Predicted Pose on the YCB-Video and MP6D Datasets}
\begin{figure*}
    \centering
    \resizebox{2\columnwidth}{!}{
    \includegraphics{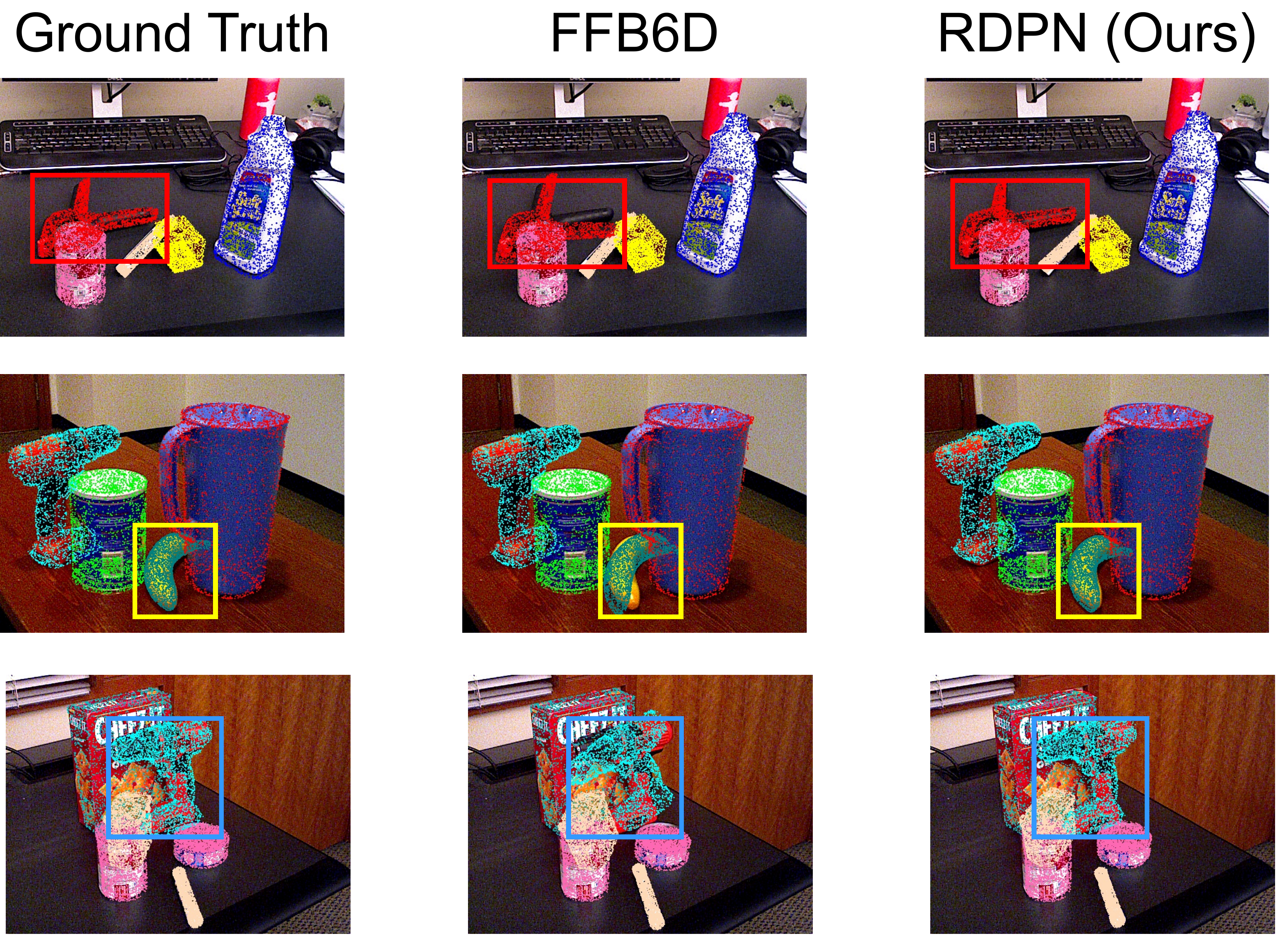}
    }
    \caption{\textbf{Qualitative results on YCB-Video dataset.} %Objects within bounding boxes demonstrate pose estimations where our method significantly outperforms the previous state-of-the-art. All images are rendered by projecting the 3D object model onto the image plane using the estimated pose. Our dense correspondence method outperforms the keypoint-based method FFB6D \cite{he2021ffb6d} in handling pose estimation under occlusion conditions.
    The first column shows the ground truth pose. The second column shows the pose estimated using the keypoint-based method FFB6D \cite{he2021ffb6d}. The third column shows the pose estimated using our RDPN approach. Inside the bounding box, we see that our dense correspondence method outperforms the keypoint-based method FFB6D \cite{he2021ffb6d} in handling pose estimation under occlusion conditions.
    }
    \label{fig:more ycbv}
\end{figure*}

\begin{figure*}
    \centering
    \resizebox{2\columnwidth}{!}{
    \includegraphics{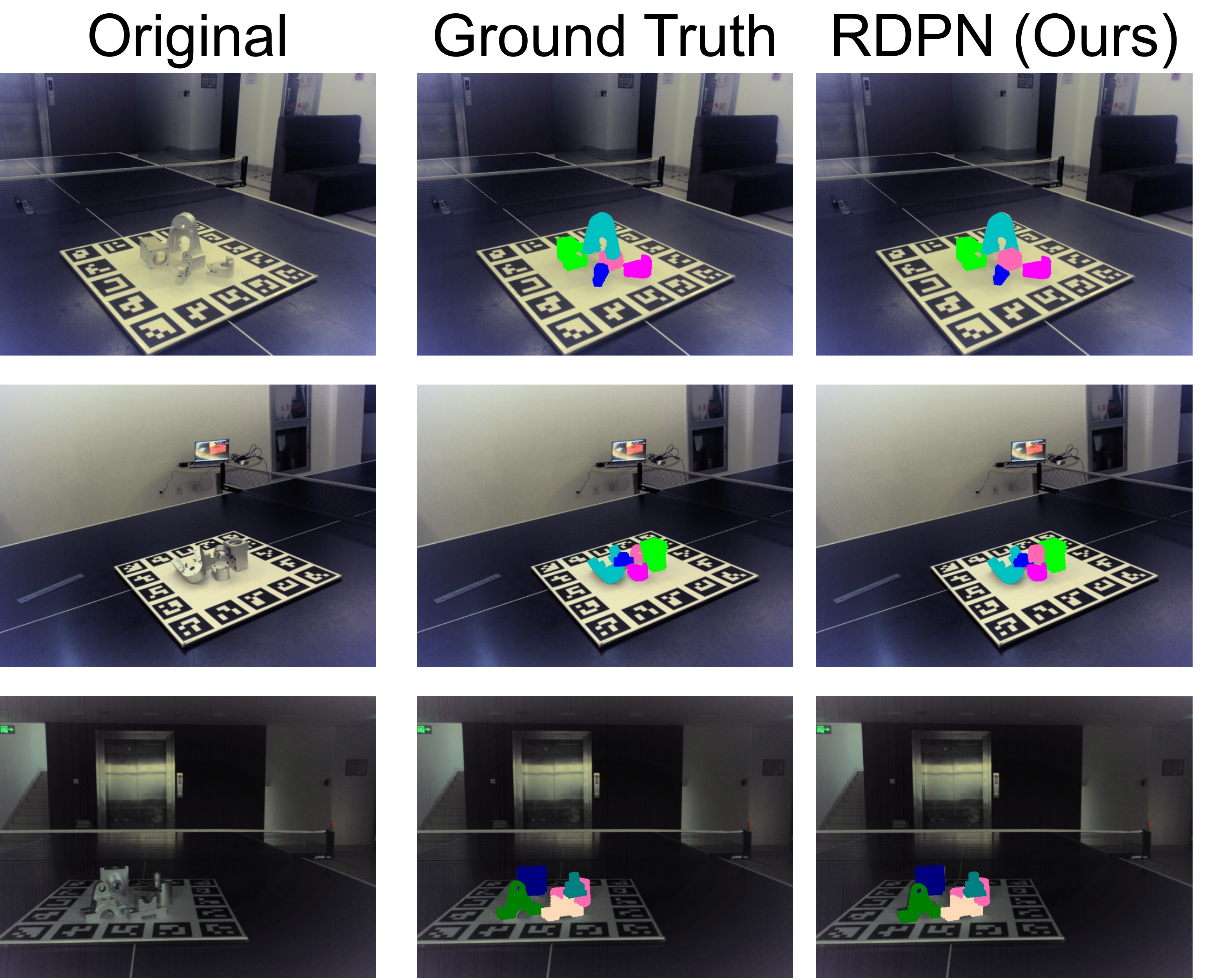}
    }
    \caption{\textbf{The qualitative results on MP6D dataset.} All images are rendered by projecting the 3D object model onto the image plane using the estimated pose. The results demonstrate the effectiveness of our method on texture-less and high-reflectivity objects under various lighting conditions.} 
    \label{fig:more mp6d}
\end{figure*}

We provide several qualitative comparison results between our method and the previous state-of-the-art method FFB6D \cite{he2021ffb6d} in \cref{fig:more ycbv} for the YCB-Video dataset. Additionally, we provide several qualitative results on the MP6D dataset in \cref{fig:more mp6d}. 

The results demonstrate the effectiveness of our method on both datasets, including \textit{texture-less} and \textit{high-reflectivity} objects.

%\label{sec:intro}

\newpage

%%%%%%%%% REFERENCES
{\small
\bibliographystyle{ieee_fullname}
\bibliography{egbib}
}

